\documentclass[10pt,twocolumn,letterpaper]{article}

\usepackage{cvpr}
\usepackage{times}
\usepackage{epsfig}
\usepackage{graphicx, subcaption}
\usepackage{amssymb}
\usepackage{xcolor}
\usepackage{booktabs}
\usepackage{makecell}
\usepackage{enumitem}% http://ctan.org/pkg/enumitem
\usepackage[accsupp]{axessibility}
\usepackage{afterpage}
\usepackage{multirow}
\usepackage{array}
\usepackage{adjustbox}
\usepackage{float}
\usepackage{graphicx}
\usepackage{lipsum}
\usepackage[pagebackref,breaklinks,colorlinks]{hyperref}

\begin{document}

\title{ Synth-SONAR: Sonar Image Synthesis with Enhanced Diversity and Realism\\ via Dual Diffusion Models and GPT Prompting }

\author{
Purushothaman Natarajan\textsuperscript{*}, Kamal Basha, Athira Nambiar\\
Department of Computational Intelligence,\\
SRM Institute of Science and Technology,\\
Kattankulathur, Tamil Nadu, 603203, India\\
{\tt\small c30945@srmist.edu.in, c58527@srmist.edu.in, athiram@srmist.edu.in}\\
}

\maketitle

\begin{abstract}

Sonar image synthesis is crucial for advancing applications in underwater exploration, marine biology, and defence. Traditional methods often rely on extensive and costly data collection using sonar sensors, jeopardizing data quality and diversity. To overcome these limitations, this study proposes a new sonar image synthesis framework, ``Synth-SONAR" leveraging diffusion models and GPT prompting. The key novelties of Synth-SONAR are threefold: First, by integrating Generative AI-based style injection techniques along with publicly available real/ simulated data, thereby producing one of the largest sonar data corpus for sonar research. Second, a dual text-conditioning sonar diffusion model hierarchy synthesizes coarse and fine-grained sonar images with enhanced quality and diversity. Third, high-level (coarse) and low-level (detailed) text-based sonar generation methods leverage advanced semantic information available in visual language models (VLMs) and GPT-prompting. During inference, the method generates diverse and realistic sonar images from textual prompts, bridging the gap between textual descriptions and sonar image generation. This marks the application of GPT-prompting in sonar imagery for the first time to the best of our knowledge. Synth-SONAR achieves state-of-the-art results in producing high-quality synthetic sonar datasets, significantly enhancing their diversity and realism. 

\end{abstract}
\vspace{-.3cm}

\section{Introduction}
\label{sec:intro}
Sound Navigation and Ranging (SONAR) technology is an essential component of underwater exploration and object detection, with extensive applications in areas such as anti-submarine warfare, mine detection, submarine navigation, and torpedo guidance. It serves both civilian and military purposes, playing a crucial role in ensuring safety and operational efficiency in challenging underwater environments. Sonar operates by emitting sound waves that travel through water, reflect off objects, and are analyzed upon return to determine the location, size, and shape of underwater objects~\cite{yang2020underwater}.

Sonar images are complex, consisting of the target, target shadow, and reverberation background regions~\cite{mignotte1999three}. Further, the underwater environment adds challenges like turbulence, noise, and low resolution, making underwater image analysis practically difficult~\cite{almutiry2024underwater}. Publicly available sonar datasets~\cite{huo2020underwater,sethuraman2024machine,zhang2021self} 
often face challenges like low resolution, poor feature representation, and limited object diversity. Their scarcity is worsened by the need for expert labeling, security issues, and data sensitivity. To overcome these limitations, simulation-based studies~\cite{s2024s3simulatorbenchmarkingscansonar,shin2022synthetic,lee2018deep} have been explored in the literature. However, these approaches still face limitations, including time-consuming manual modeling, the complexity of integrating various tools, and insufficient diversity in generated sonar data. 

To this end, machine learning (ML)/ deep learning (DL) techniques have been employed for efficient sonar image synthesis~\cite{jegorova2020full,yang2024sample,koo2024cycle,jiang2020side} in the recent years. Building on the success of generative AI (GenAI) in various fields e.g. medical imaging~\cite{shokrollahi2023comprehensive,nie2017medical}, autonomous vehicles~\cite{yu2018intelligent} etc. similar approaches such as Generative Adversarial Networks (GANs) ~\cite{goodfellow2014generative}, style transfer ~\cite{gatys2015neural} and Denoising Diffusion Probabilistic Models (DDPMs) ~\cite{ho2020denoising} have been adopted in sonar image synthesis to tackle data scarcity and improve model accuracy. Nonetheless, significant challenges remain, including insufficient object diversity, poor fine-grained feature preservation, complexities such as target shadows and reverberation effects, and high computational demands for real-time sonar image synthesis. Additionally, there exists a semantic gap between domain experts and machine learning models in interpreting and explaining the sonar characteristics in a human-compliant way.

In this paper, we propose a novel sonar image synthesis framework i.e. \textbf{Synth-SONAR} to overcome the aforementioned limitations. Synth-SONAR generates high-quality, realistic sonar images by dint of advanced GenAI techniques i.e. text-conditioned diffusion models and GPT promting. The workflow of Synth-SONAR consists of three phases. In the first phase,  a large-scale, diverse corpus of sonar data by integrating publicly available sonar images, S3 Simulator~\cite{s2024s3simulatorbenchmarkingscansonar}, and style-injected sonar images captioned using CLIP~\cite{radford2021learning} based vision-language models to capture high-level semantic features is generated. The next phase involves training a Denoising Diffusion Probabilistic Model (DDPM)~\cite{ho2020denoising}, fine-tuned with LoRA (Low-Rank Adaptation)~\cite{hu2021lora} and integrated with GPT~\cite{brown2020language} based prompts to generate coarse-level sonar images. Finally, in the final phase, the coarse images are refined into fine-grained outputs using domain-specific language instructions processed through a Vision-Language Model (VLM), further enhanced with LoRA fine-tuning and GPT for content precision. Our approach achieves a high degree of diversity and realism, as demonstrated through extensive qualitative and quantitative analysis, via metrics such as Fréchet Inception Distance (FID), Peak Signal-to-Noise Ratio (PSNR), Structural Similarity Index (SSIM), and Inception Score (IS). The key contributions of the paper are as follows:

\begin{itemize}
    \item A novel GenAI framework i.e.``Synth-SONAR" for sonar image synthesis, incorporating dual-stage text-conditioned diffusion models for high-quality, multi-resolution image generation.
    \vspace{-.2cm}
    \item One of the most extensive and diverse sonar image datasets through the integration of multiple sources (real, simulated, and GenAI) and detailed annotations.
    \vspace{-.4cm}
    \item An innovative approach that enhances image generation techniques by utilizing Denoising Diffusion Probabilistic Models (DDPM) combined with LoRA and GPT-based prompts for controlled and high-quality realistic sonar image synthesis, thus making our approach interpretable.
\end{itemize}

\section{Related Work}
\label{sec: related works}

\subsection{Underwater Sonar image analysis}
Earlier sonar image analysis used traditional Machine learning (ML) techniques such as the Markov random field (MRF) model with the scale causal multigrid (SCM) algorithm~\cite{847834} and undecimated discrete wavelet transform (UDWT) combined with PCA and k-means clustering~\cite{5732666}. Advancements in deep learning (DL) introduced methods like FS-UTNet~\cite{10531031}, a framework for underwater target detection using few-shot learning. Further, techniques such as RotNet, Denoising Autoencoders, and Jigsaw~\cite{Preciado-Grijalva_2022_CVPR} facilitated learning representations for sonar image classification without large labeled datasets. EsonarNet~\cite{10614610}, a lightweight vision transformer network, is designed for efficient segmentation. The Global Context External-Attention Network (GCEANet) provides zero-shot classification in \cite{10399359}. YOLOv7 improves high-precision object detection by integrating Swin-Transformer and Convolutional Block Attention Module (CBAM)\cite{10534346}. EfficientNet is used as a backbone for feature extraction in ~\cite{arjun2024unveiling}, which uses dual-channel attention mechanisms (SE and ECA) and a modified BiFPN for multi-scale feature fusion. Additionally, DSA-Net for underwater object detection ~\cite{10400199} used a dual spatial attention network (DSAM), and Generalized Focal Loss (GFL) for optimized object detection. To enhance interpretability, LIME and SP-LIME have been employed~\cite{natarajan2024underwater} to make sonar image classification more transparent and understandable.

\begin{figure*}[t!]
    \centering
    \includegraphics[width=\linewidth]{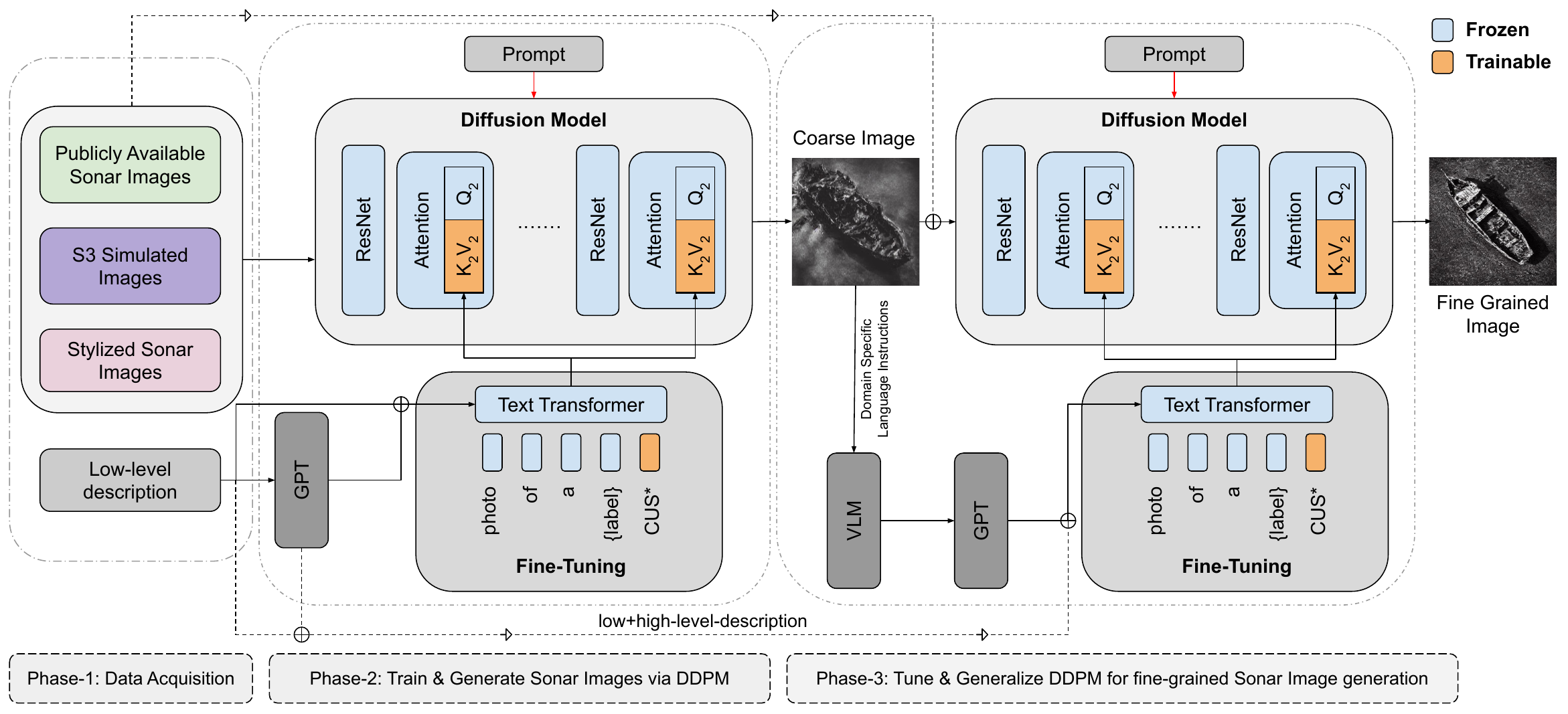}
    \caption{Overall Architecture of the proposed ``Synth-SONAR" sonar image synthesis framework.}
    \vspace{-.3cm}
    \label{fig:overall architecture}
\end{figure*}

\subsection{Synthetic Data Generation via Generative AI}
Synthetic data generation via GenAI addresses data scarcity using generative AI techniques like diffusion models~\cite{cao2024survey}, GANs~\cite{goodfellow2020generative}, and VAEs~\cite{kingma2013auto}. CAD-based methods include using Unreal Engine (UE)~\cite{shin2022synthetic} to create sonar images with diverse seabed conditions and objects, and the S3Simulator dataset~\cite{s2024s3simulatorbenchmarkingscansonar}, which leverages advanced simulation techniques, Segment Anything Model (SAM), and tools like SelfCAD and Gazebo for 3D modeling. Augmentation techniques feature the Seg2Sonar network~\cite{10440265}, which uses spatially adaptive denormalization (SPADE), Skip-Layer channel-wise Excitation (SLE), and weight adjustment (WA) modules. GAN-based methods include an enhanced CycleGAN~\cite{10472044} model that improves underwater image contrast through a depth-oriented attention mechanism, the CBL-sinGAN network~\cite{peng2024sample} which combines sinGAN with Convolutional Block Attention Module (CBAM) for target image augmentation. Further, SIGAN~\cite{peng2024sigan} used a multi-scale GAN for super-resolution of sonar images. Diffusion-based approaches are exemplified by a method that uses diffusion models for synthetic image generation and augmentation in ~\cite{10453275,yang2023side} and an enhanced YOLOv7 model that integrates a denoising-diffusion model~\cite{10534346},  Vision Transformer (ViT) for high-precision object detection in side-scan sonar images.

\subsection{Vision-Language Models (VLMs) and Their Application in the Sonar Domain}
Vision-Language Models (VLMs) represent a significant advancement in the field of artificial intelligence by bridging the gap between visual and textual data~\cite{zhang2024vision}.
Multimodels like CLIP~\cite{radford2021learning}, DALL-E~\cite{Ramesh2021ZeroShotTG}, BLIP~\cite{li2022blip}, FLAVA ~\cite{singh2022flava}, and GIT ~\cite{parmar2018image} represent significant advancements by integrating visual and textual data, enhancing image generation and understanding through multimodal learning. While VLMs have made strides in various domains, their application to sonar data is still in its nascent stage, with only very few works in the sonar domain e.g. VALE~\cite{natarajan2024valemultimodalvisuallanguage}, which combines VLM techniques with sonar data for improved underwater environment analysis.

\section{Methodology}
\label{sec: methodology}

In this section, we provide a detailed description of the proposed Synth-SONAR framework for generating sonar images. The overall architecture of the model is depicted in Fig.~\ref{fig:overall architecture}, which consists of three key phases. Phase 1 is the \textbf{Data Acquisition Phase}, as explained in Section \ref{subsec: data acquisition}. It entails the collection of real-world, CAD-simulated, and Gen-AI-generated images. Phases 2 and 3 utilize \textbf{text-conditioned dual diffusion models and GPT-based prompting} to synthesize both ``coarse" and ``fine" grained sonar images. In particular, Phase-2 customizes pre-trained diffusion models for generating sonar images as described in Section \ref{subsec: customizing diffusion models}. Whereas, Phase-3 fine-tunes and generalizes the diffusion models for generating ``fine"-grained sonar images as described in Section \ref{subsec: generalizing diffusion models}.

\subsection{Phase-1: Data Acquisition}
\label{subsec: data acquisition}
 
Underwater sonar imagery is a critical domain, wherein the data collection and processing of such a large training dataset is both expensive and challenging. Some of the common ways are to leverage the publicly available datasets e.g. Seabed Objects KLSG dataset, SCTD (see Section \ref{subsubsec: publicly available dataset}) or the CAD-based simulated dataset such as S3 simulator data (see Section \ref{subsubsec: s3 simular dataset}). Further, we advance the sonar image synthesis via Generative AI techniques such as style injection (see Section \ref{subsubsec: style injection}), as explained in the forthcoming section.

\subsubsection{Style Injection on Generated Images}
\label{subsubsec: style injection}

\begin{figure}[t!]
    \centering
    \includegraphics[width=\linewidth]{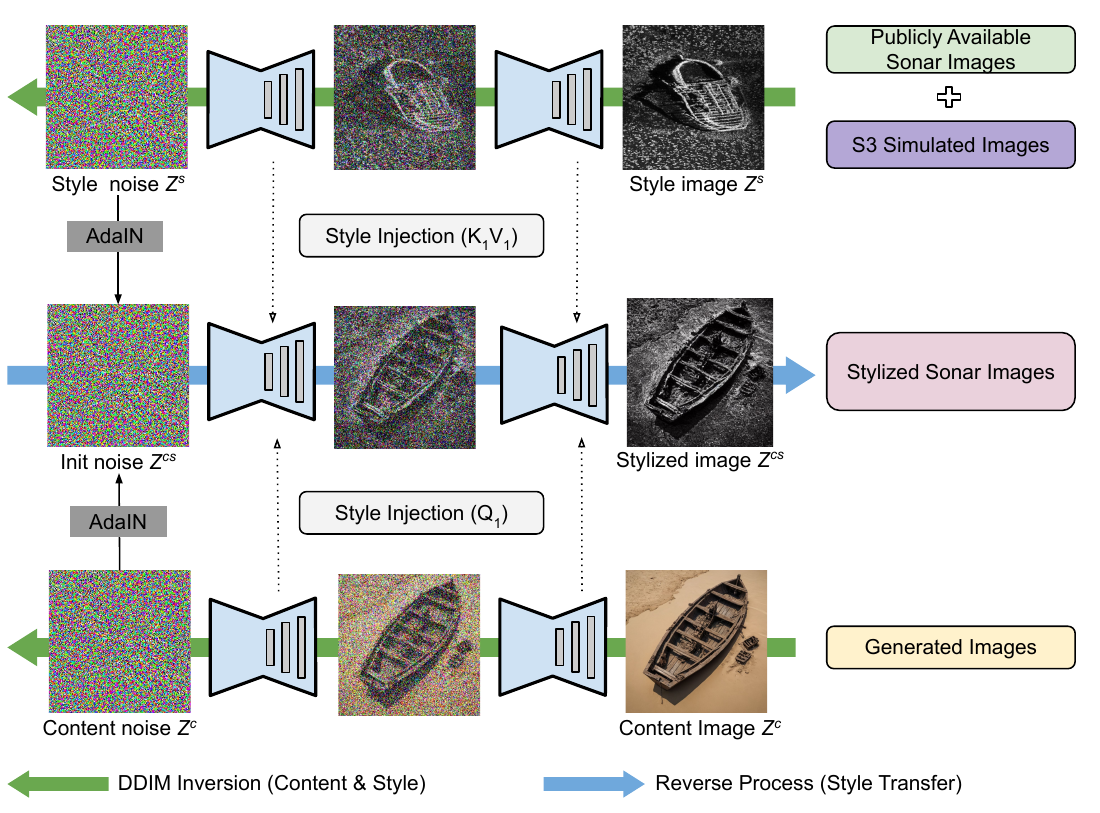}
    \caption{Style Injection for sonar Image Synthesis.}
    \label{fig: data acquisition-2}
    \vspace{-.5cm}
\end{figure}

Style injection is a technique in computer vision's image-to-image tasks, that combines content and style features to generate a new image, where the objective is to transform the content image (in our case, it is generated by the prompts from GPT) by infusing it with the stylistic elements of another image while preserving the original structure~\cite{chung2024style}. In this work, style injection is utilized to add more diversity to the real sonar data and to increase the availability of sonar data. Refering to Fig.~\ref{fig: data acquisition-2}, we leverage the generative capability of a pre-trained large-scale model to generate content images and transfer sonar style information to the generated content images using a pre-trained stable diffusion model, thereby resolving the issue of the traditional data-collection process.

We employ Latent Diffusion Model (LDM) \cite{rombach2022high} to perform image synthesis by operating within a low-dimensional latent space. This approach significantly reduces computational costs while maintaining a focus on the semantic content of the data. For a given image \( x \in \mathbb{R}^{H \times W \times 3} \), the encoder \( E \) encodes the image into a latent representation \( z \in \mathbb{R}^{h \times w \times c} \), and the decoder reconstructs the image from this latent representation. The diffusion model is trained on the latent space \( z \), where the task is to predict noise \( \epsilon \) from the noised latent representation \( z_t \) at a given time step \( t \). The corresponding training objective is:

\begin{equation}
    L_{LDM} = \mathbb{E}_{z, \epsilon, t}\left[\| \epsilon - \epsilon_\theta(z_t, t, y) \|_2^2 \right]
\end{equation}

where \( \epsilon \sim \mathcal{N}(0, 1) \) is the noise, \( t \) is uniformly sampled from \(\{1, \ldots, T\}\), \( y \) is the conditioning variable (which could be a style reference or text prompt), and \( \epsilon_\theta \) is a neural network predicting the noise added to \( z \).

\textbf{Generic Attention Mechanism:}
In Latent Diffusion Models (LDM), the attention mechanism is fundamental to both style injection in image-to-image tasks and text-to-image generation. The attention mechanism can be generically expressed as:

\begin{equation}
    Q = W_Q(\phi), \quad K = W_K(\psi), \quad V = W_V(\psi)
    \label{equ: generic attension mechanism-1}
\end{equation}

\begin{equation}
    \phi_{out} = \text{Attn}(Q, K, V) = \text{softmax}\left(\frac{QK^T}{\sqrt{d}}\right) \cdot V
    \label{equ: generic attension mechanism-2}
\end{equation}

where, query \(Q\), key \(K\), and value \(V\) are the result of learned linear transformations via the projection layers \( W_Q(\cdot) \), \( W_K(\cdot) \), and \( W_V(\cdot) \), respectively. The dimensionality of these projections is denoted by \(d\), which is used in scaling the dot product between \(Q\) and \(K\) in the attention mechanism. Also,
\( \phi \) represents the feature output from the residual block, while \( \psi \) is the conditioning variable, which can be the style features for image-to-image tasks or the text features for text-to-image tasks.  

\textbf{Style Injection in Image-to-Image Tasks:}
For Style Injection, the conditioning variable \( \psi \) corresponds to the style features extracted from a reference image. The goal of style injection is to blend the style features with the content features of the original image, transferring the style while maintaining the content's structure. The content features are represented as:

\begin{equation}
    Q_c = W_Q(\phi)
\end{equation}

where \( \phi \) denotes the content features extracted from the image. The style features are extracted from the style image and projected into key and value pairs:

\begin{equation}
    K_s = W_K(\phi_s), \quad V_s = W_V(\phi_s)
\end{equation}

where \( \phi_s \) represents the style features from the reference image. The attention mechanism used for injecting the style features into the content is given by:

\begin{equation}
    \tilde{Q}^{cs}_t = \gamma \times Q^c_t + (1 - \gamma) \times Q^{cs}_t
    \label{equ: style injection}
\end{equation}

\begin{equation}
    \phi^{cs}_{out} = \text{Attn}(\tilde{Q}^{cs}_t, K^{s}_t, V^{s}_t)
\end{equation}

where \( \gamma \) is a blending ratio that controls the amount of style injected. For sonar-specific style injection, referring to Fig.~\ref{fig: data acquisition-2}, the latent noise representations of the content and style images are denoted by \( z^c \) and \( z^s \), respectively. These representations are obtained through the DDIM inversion process, which reconstructs both the content and style images into Gaussian noise at time step \( t = T \). During the DDIM inversion process, we collect the query features \( Q^c_t \) from the content and the key-value pairs \( K^s_t \), \( V^s_t \) from the style at each time step \( t \). Once the inversion is completed, we initialize the stylized latent noise \( z^{cs}_T \) by copying the content latent noise \( z^c_T \). To inject the style into the content, we blend the content and stylized queries using a blending ratio \( \gamma \) from equation~\ref{equ: style injection}. The blended query \( \tilde{Q}^{cs}_t \) is then passed through the attention mechanism along with the style's key and value features. By adjusting \( \gamma \), we control the degree of style transfer, where a higher \( \gamma \) retains more of the original content features, and a lower \( \gamma \) strengthens the influence of the style features. This approach allows for precise control over the stylistic outcome, ensuring a smooth transition between content preservation and style injection.

\subsection{Phase-2: Train and Generate sonar Images via DDPM and GPT-prompting}
\label{subsec: customizing diffusion models}

Our objective is to generate a sequence of sonar images that correspond to the given text conditioning. In the case of text-to-image generation, referring to the equation~\ref{equ: generic attension mechanism-1}, the conditioning variable \( \psi \) corresponds to the text features. The cross-attention mechanism aligns the textual description with the visual features to generate an image that matches the text prompt.

$Q_2$, $K_2$, and $V_2$ are analogous to the query, key, and value in the diffusion model explained earlier but are specific to \textit{Phase 2}. In Phase 2, these components represent the following: \( Q_2 = W_Q(\phi) \) represents the \textit{image features}, \( K_2 = W_K(\text{Text}) \), and \( V_2 = W_V(\text{Text}) \) represent the \textit{text features}. The attention mechanism for text-to-image generation is given by:

\begin{equation}
    \phi^{out} = \text{Attn}(Q_2, K_2, V_2) = \text{softmax}\left(\frac{Q_2 K_2^T}{\sqrt{d}}\right) \cdot V_2
\end{equation}

where \( \phi^{out} \) represents the output of the cross-attention layer, combining the image features with the text features. This interaction between the image and text features enables the generation of images conditioned on the provided text prompt.

To train a diffusion model from scratch, enormous computation and a vast amount of training data are required. To mitigate the challenges of data scarcity and computation, we adopted LoRA~\cite{hu2021lora} for fine-tuning. LoRA (Low-Rank Adaptation) applies to text-to-image tasks by efficiently fine-tuning models. Mathematically, LoRA updates the model's weight matrices by adding a low-rank decomposition \( \Delta W = A \cdot B \), where \( A \in \mathbb{R}^{m \times r} \) and \( B \in \mathbb{R}^{r \times n} \) are low-rank matrices, where \( r \) is the rank.

LoRA updates the query, key, and value matrices in the following manner:

\begin{equation}
    Q_2' = Q_2 + \alpha \cdot \Delta W_{Q_2}
\end{equation}

\begin{equation}
    K_2' = K_2 + \alpha \cdot \Delta W_{K_2}
\end{equation}

\begin{equation}
    V_2' = V_2 + \alpha \cdot \Delta W_{V_2}
\end{equation}

where, \( Q_2', K_2', V_2' \) represent the LoRA-enhanced \textbf{queries}, \textbf{keys}, and \textbf{values}, where the original \( Q_2 \), \( K_2 \), and \( V_2 \) are adjusted by adding the LoRA updates \( \Delta W_{Q_2}, \Delta W_{K_2}, \Delta W_{V_2} \), scaled by a factor \( \alpha \). The enhanced attention mechanism with LoRA is given by:

\begin{equation}
    \phi^{c}_{out} = \text{Attn}(Q_2', K_2', V_2') = \text{softmax}\left(\frac{Q_2' \cdot K_2'^T}{\sqrt{d_2}}\right) \cdot V_2'
\end{equation}

The attention mechanism, \( \text{Attn}(Q_2', K_2', V_2') \), now uses the LoRA-enhanced queries, keys, and values to generate the final output \( \phi^{c}_{out} \) in text-to-image tasks.

\subsection{Phase-3: Tune \& Generalize DDPM for Fine-Grained sonar Image Generation}
\label{subsec: generalizing diffusion models}

The fine-tuned model from Phase 2 can generate images only when the user provides a custom object tag given during training. If the user fails to provide this tag, the model generates regular images instead of SONAR-specific images. To mitigate this limitation and generalize the model for generating fine-grained sonar images, we further train the model using the existing corpus of sonar images.

In this phase, we generate images from the Phase 2 model using a series of prompts obtained from GPT. These generated images, combined with domain-specific language instructions, are processed through a Visual Language Model (VLM) to obtain \textbf{low-level descriptions}. These low-level descriptions are then fed back into GPT to generate \textbf{high-level descriptions}. The images and their corresponding low-level + high-level descriptions are subsequently used to fine-tune the stable diffusion model. The fine-tuning process involves optimizing the loss function for the diffusion model, where the loss is conditioned on both the image data and the text prompt:

\begin{equation}
    \mathcal{L}_{\text{fine-tune}} = \mathbb{E}_{x, t, \epsilon} \left[ \left\| \epsilon - \epsilon_{\theta}(x_t, t | \text{prompt}) \right\|^2 \right]
\end{equation}

Here, \( x_t \) represents the generated images, \( \epsilon_{\theta} \) is the noise predictor, and the fine-tuning process incorporates both the image data and sonar domain-specific textual instructions. This enhances the model's ability to generate fine-grained sonar images based on more generalized prompts.

\section{Experimental Setup}
\label{sec:experiments}

\subsection{Dataset}
\label{subsec: dataset}

\subsubsection{Publicly Available Sonar Images}
\label{subsubsec: publicly available dataset}
\begin{figure}[h!]
    \centering
    \begin{minipage}{0.22\columnwidth}
        \centering
        \fontsize{8}{10}\selectfont\text{Seafloor} % Label for the first image
    \end{minipage}
    \begin{minipage}{0.22\columnwidth}
        \centering
        \fontsize{8}{10}\selectfont\text{Plane} % Label for the second image
    \end{minipage}
    \begin{minipage}{0.22\columnwidth}
        \centering
        \fontsize{8}{10}\selectfont\text{Ship} % Label for the third image
    \end{minipage}
    \begin{minipage}{0.22\columnwidth}
        \centering
        \fontsize{8}{10}\selectfont\text{Human} % Label for the fourth image
    \end{minipage}
    \vspace{0.05cm}
    \begin{minipage}{0.22\columnwidth}
        \centering
        \includegraphics[width=\linewidth]{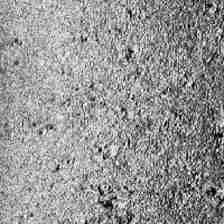}
    \end{minipage}
    \begin{minipage}{0.22\columnwidth}
        \centering
        \includegraphics[width=\linewidth]{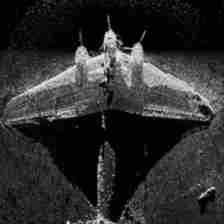}
    \end{minipage}
    \begin{minipage}{0.22\columnwidth}
        \centering
        \includegraphics[width=\linewidth]{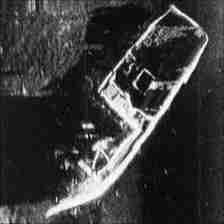}
    \end{minipage}
    \begin{minipage}{0.22\columnwidth}
        \centering
        \includegraphics[width=\linewidth]{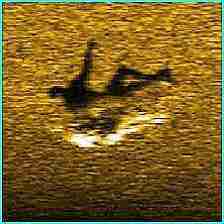}
    \end{minipage}
    \caption{Samples from publicly available sonar images}
    \label{fig:publicly_available_sonar_images-act}
\end{figure}

In this work, publicly available datasets, specifically the Seabed Objects KLSG~\cite{huo2020underwater} and the sonar Common Target Detection Dataset (SCTD)~\cite{zhang2021self}, are utilized. The Seabed Objects KLSG dataset ~\cite{huo2020underwater} includes 1,190 side-scan sonar images, featuring 385 shipwrecks, 36 drowning victims, 62 planes, 129 mines, and 578 seafloor images. Collected over ten years with the help of commercial sonar suppliers like Lcocean, Klein Martin, and EdgeTech, a subset of 1,171 images (excluding mines) is publicly available for academic research. The sonar Common Target Detection Dataset (SCTD) 1.0 ~\cite{zhang2021self}, developed by multiple universities, contains 596 images across 3 classes. Some sample images of the publicly available data are shown in Fig.~\ref{fig:publicly_available_sonar_images-act}.

\vspace{-.3cm}
\subsubsection{S3 Simulator Images}
\label{subsubsec: s3 simular dataset}

\begin{figure}[h!]
\vspace{-.25cm}
    \centering
    \begin{minipage}{0.20\columnwidth}
        \centering
        \fontsize{8}{10}\selectfont\text{Ship} % Adjust the size here
    \end{minipage}
    \begin{minipage}{0.20\columnwidth}
        \centering
        \fontsize{8}{10}\selectfont\text{Plane} % Adjust the size here
    \end{minipage}
    \begin{minipage}{0.20\columnwidth}
        \centering
        \fontsize{8}{10}\selectfont\text{Manta Mine} % Adjust the size here
    \end{minipage}
    \begin{minipage}{0.20\columnwidth}
        \centering
        \fontsize{8}{10}\selectfont\text{Cylindrical Mine} % Adjust the size here
    \end{minipage}
    \vspace{0.05cm}

    \centering
    \begin{minipage}{0.20\columnwidth}
        \centering
        \includegraphics[width=\linewidth]{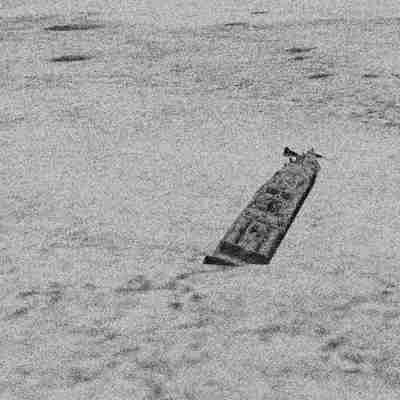}
    \end{minipage}
    %\hspace{0.0001\columnwidth}
    \begin{minipage}{0.20\columnwidth}
        \centering
        \includegraphics[width=\linewidth]{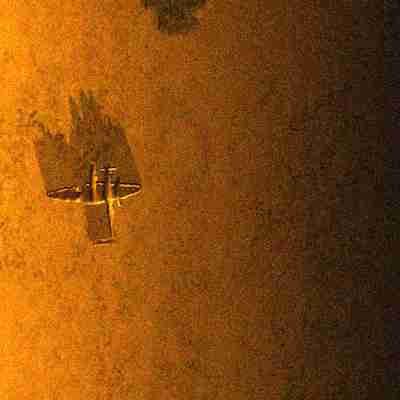}
    \end{minipage}
    %\hspace{0.0001\columnwidth}
    \begin{minipage}{0.20\columnwidth}
        \centering
        \includegraphics[width=\linewidth]{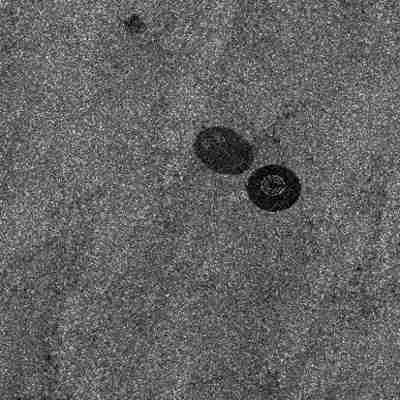}
    \end{minipage}
    %\hspace{0.001\columnwidth}
    \begin{minipage}{0.20\columnwidth}
        \centering
        \includegraphics[width=\linewidth]{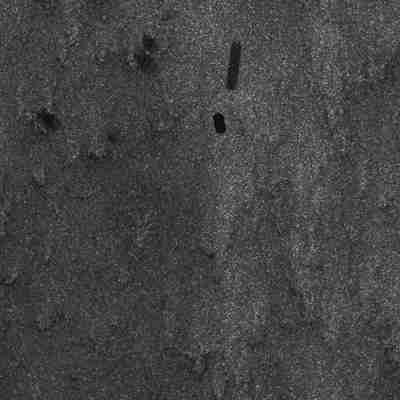}
    \end{minipage}
    \vspace{0.1cm}
    \caption{Samples from S3 Simulator Dataset}
    \label{fig: S3SIM}
\end{figure}

The S3Simulator~\cite{s2024s3simulatorbenchmarkingscansonar} dataset is a novel benchmark of simulated side-scan sonar images designed to overcome challenges in acquiring high-quality sonar data. Using advanced simulation techniques, the dataset accurately replicates underwater conditions and produces diverse synthetic sonar images, as illustrated in Fig.~\ref{fig: S3SIM}. Tools like the Segment Anything Model (SAM) and Gazebo are utilized for optimal object segmentation and visualization, enhancing the quality of data for AI model training in underwater object classification.

\subsection{Evaluation Metrics}
\label{Subsec: evaluation metrics}
We use Frechet Inception Distance (FID), Structural Similarity Index (SSIM), Peak Signal-to-Noise Ratio (PSNR), and Inception Score (IS) to evaluate the image-to-image generation task.

\begin{equation}
    \text{FID}(x, g) = ||\mu_x - \mu_g||^2 + \text{Tr}(\Sigma_x + \Sigma_g - 2(\Sigma_x \Sigma_g)^{1/2})
\end{equation}

where \( \mu_x \) and \( \mu_g \) are the means, and \( \Sigma_x \) and \( \Sigma_g \) are the covariance matrices of the real images \( x \) and the generated images \( g \). The term \( \text{Tr} \) denotes the trace of the matrix. FID measures the distance between the distributions of real and generated images in the feature space of a pre-trained Inception model.

\begin{equation}
    \text{SSIM}(x, y) = \frac{(2\mu_x \mu_y + c_1)(2\sigma_{xy} + c_2)}{(\mu_x^2 + \mu_y^2 + c_1)(\sigma_x^2 + \sigma_y^2 + c_2)}
\end{equation}

where \( \mu_x \) and \( \mu_y \) represent the means of images \( x \) and \( y \), \( \sigma_x^2 \) and \( \sigma_y^2 \) are their variances, and \( \sigma_{xy} \) is the covariance between the two images. Constants \( c_1 \) and \( c_2 \) are used to avoid instability when the denominator is close to zero. SSIM assesses the perceived quality of images based on luminance, contrast, and structure.

\begin{equation}
    \text{PSNR}(x, y) = 10 \cdot \log_{10}\left(\frac{\text{MAX}^2}{\text{MSE}(x, y)}\right)
\end{equation}

where \( \text{MAX} \) is the maximum possible pixel value of the image (e.g., 255 for an 8-bit image) and \( \text{MSE}(x, y) \) is the mean squared error between the images \( x \) and \( y \). PSNR is commonly used to measure the quality of reconstruction in image compression.

\begin{equation}
    \text{IS}(x) = \exp\left(\mathbb{E}_x D_{\text{KL}}(p(y|x) || p(y))\right)
\end{equation}

where \( p(y|x) \) is the conditional label distribution given an image \( x \), and \( p(y) \) is the marginal label distribution. \( D_{\text{KL}} \) represents the Kullback-Leibler divergence. IS evaluates the quality and diversity of generated images based on how well they are classified by a pre-trained classifier.

To evaluate the performance of the text-to-image generation task, R-Precision, Multimodal Distance~\cite{srivastava2012multimodal}, and Diversity can be used. However, since the sonar dataset lacks ground truth, the text-to-image sonar image generation model cannot be evaluated quantitatively; instead, qualitative analysis is facilitated.

\subsection{Implementation Details}
\label{Subsec: implementation details}
We train, develop, and experiment the proposed framework in stable diffusion 1.4 pre-trained on the LAION~\cite{schuhmann2022laion} dataset using a NVIDIA RTX4000 with a 20GB GPU, fine-tuned with sonar-specific images to tailor the synthesis to the target sonar domain. The diffusion model generates high-quality synthetic sonar images by conditioning on sonar-specific attribute styles in the case of image-to-image and domain-specific prompts in the case of text-to-image. The domain-specific prompts (low-level and high-level descriptions) are enhanced using GPT 3.5 Turbo at phase-2 and in phase-3 VLM-LLaVA~\cite{liu2024visual} and GPT-3.5 Turbo for improving the control over sonar image synthesis. For the classification task, we leverage transfer learning with several backbone models, including VGG16, ResNet50, DenseNet121, MobileNetV2, Xception, and InceptionResNetV2. The models pre-trained on large-scale ImageNet dataset~\cite{deng2009imagenet} are adapted for sonar image classification with real dataset, synthetic dataset and a combination of real and synthetic dataset.

\section{Experimental Results}
\label{sec: results}

\subsection{Phase-1: Style-Injection Results}
\label{subsec: image-to-image}

\subsubsection{Style-Injection Quantitative Results}
\label{subsubsec: style injection quantitative results}
The objective of style injection is to improve the quality and diversity of generated sonar images through style injection, ensuring that the generated images better reflect the unique characteristics of the dataset. To achieve this, style images are selected not randomly but via K-Means clustering on the real sonar dataset. One to three images from each cluster are chosen for style injection, which enhances the relevance and consistency of the stylized outputs.

\begin{table}[h!]
    \centering
    \begin{tabular}{|c|c|c|c|c|}
        \hline
        \textbf{Class} & \textbf{FID} & \textbf{SSIM} & \textbf{PSNR} & \textbf{IS} \\
        \hline
        Plane & 4.13 & 0.300 & 15.022 & 1.05 \\ 
        Ship & 6.15 & 0.392 & 10.759 & 1.04 \\  
        Seafloor & 1.12 & 0.452 & 12.410 & 1.12 \\
        \hline
        Average & 3.8 & 0.381 & 12.730 & 1.07 \\  
        \hline 
    \end{tabular}
    \vspace{.1cm}
    \caption{Image-to-Image Quantitative Metrics}
    \label{tab: Image-to-image evaluation}
\end{table}

The generated images from Phase-1: Style Injection (Section 5.1.2) are quantitatively evaluated using FID, SSIM, PSNR, and IS scores, as shown in Table \ref{tab: Image-to-image evaluation}. Lower FID scores and higher SSIM and PSNR values indicate that the style injection method successfully maintains image quality, while the IS scores reflect the diversity of the generated images. From the results, the FID score is lowest for the Seafloor class (1.12), indicating the highest fidelity to the real data, while the Plane class has the lowest PSNR (15.022), suggesting relatively lower noise in image reconstruction. The SSIM values show that structural similarity is strongest for the Seafloor class (0.452), and the average across all classes indicates that the style injection improves both image quality and consistency. The IS scores across all classes are fairly similar, but they still indicate a moderate level of diversity in the generated images. These results suggest that improvements in both image quality and diversity are due to effective style injection, which introduces stylistic variations while preserving key structural elements.

\begin{table*}[t!]
    \centering
    \begin{tabular}{|m{0.12\textwidth}|m{0.35\textwidth}|m{0.2\textwidth}|m{0.2\textwidth}|}
    \hline
    \textbf{Image} & \textbf{Prompt} & \textbf{Low-Level Description} & \textbf{High-Level Description} \\ 
    \hline
    \includegraphics[width=2.2cm]{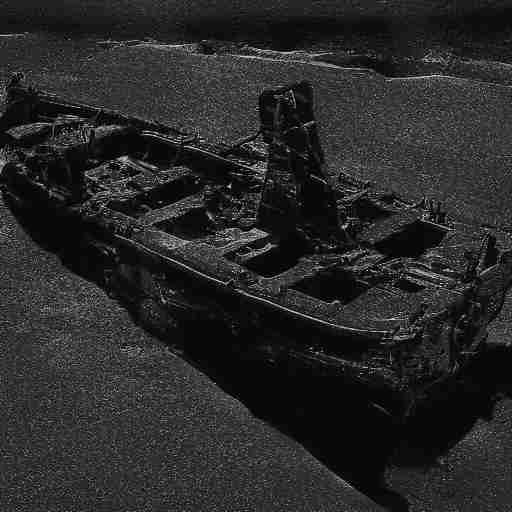} & 
    \tiny{
    Given the side scan sonar image and the caption: "image of SH\_\_* on the AP\_\_*", where PL\_\_*, SH\_\_*, CYM*, ASF*, TCM* represent the objects in the image, and AS\_\_*, AP\_\_*, SEF* represent the background. The numbers following these abbreviations can range from one to five digits. Provide the following descriptions:
    A low-level description focusing on simple details and objects visible in the image.
    A high-level description interpreting the scene or conveying a broader understanding based on the image and the given caption.
    } & \tiny
    {The image shows an old shipwreck that has sunk into water near land, with the tail end pointing upwards. The sides are positioned halfway along the ship's body, with the structure slightly off-center both horizontally and vertically. However, the ship is still identifiable as a specific model type due to the SH33* designation.} & \tiny{
    The image likely depicts a shipwreck, designated SH33*, resting on a seabed with the AP637* terrain as the background. The tail of the ship sticks out, indicating a wreck.} \\ 
    \hline
    \end{tabular}
    \vspace{0.15cm}
    \caption{\textcolor{black}{Descriptions generated by the GPT in Phase 3. In this phase, images produced from Phase 2 are described using a domain-specific prompt. The table includes four columns: (1) \textbf{Image}: the generated image from Phase 2, (2) \textbf{Prompt}: the domain-specific prompt used to generate the descriptions, (3) \textbf{Low-Level Description}: a detailed description focusing on simple, observable details and objects in the image, and (4) \textbf{High-Level Description}: an interpretive description providing a broader understanding of the scene depicted in the image. The abbreviations used in the prompt (e.g., PL\_\_*, SH\_\_*, CYM*, ASF*, TCM*, AS\_\_*, AP\_\_*, SEF*) represent various objects and background elements in the image, with numbers indicating different instances.}}
    \label{table:descriptions_from_the_llm}
\end{table*}

\subsubsection{Style-Injection Qualitative Results}
\label{subsec: style-injection qualitative results}

\begin{figure}[h!]
    \centering
    \begin{minipage}{0.25\columnwidth}
        \centering
        \textbf{CONTENT}
    \end{minipage}
    \begin{minipage}{0.25\columnwidth}
        \centering
        \textbf{STYLE} 
    \end{minipage}
    \begin{minipage}{0.25\columnwidth}
        \centering
        \textbf{OUTPUT} 
    \end{minipage}
    \vspace{0.15cm}

    % First row of images
    \begin{minipage}{0.25\columnwidth}
        \centering
        \includegraphics[width=\linewidth]{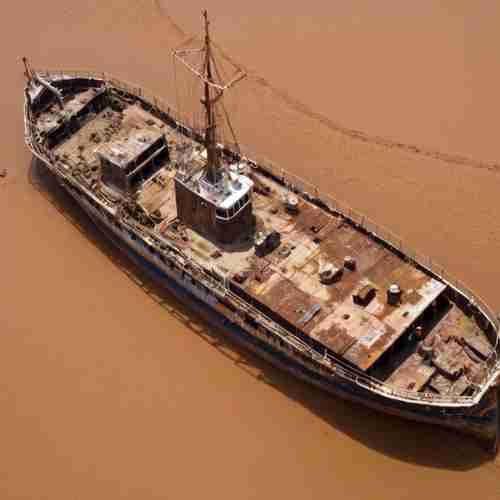}
    \end{minipage}
    \begin{minipage}{0.25\columnwidth}
        \centering
        \includegraphics[width=\linewidth]{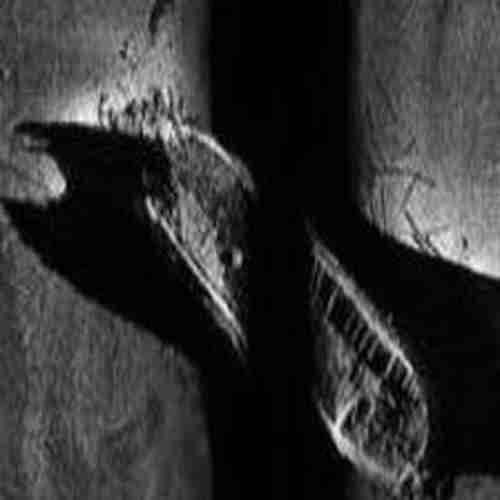}
    \end{minipage}
    \begin{minipage}{0.25\columnwidth}
        \centering
        \includegraphics[width=\linewidth]{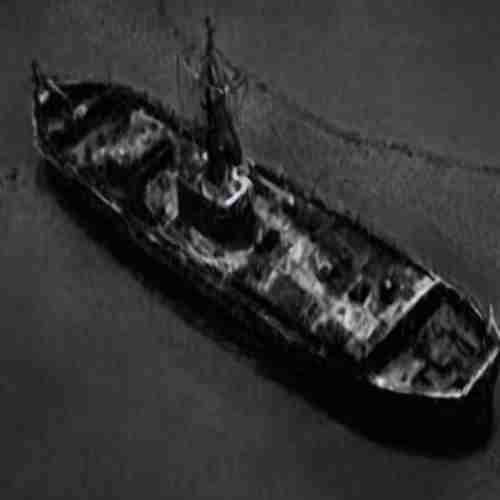}
    \end{minipage}
    % Third row of images
    \begin{minipage}{0.25\columnwidth}
        \centering
        \includegraphics[width=\linewidth]{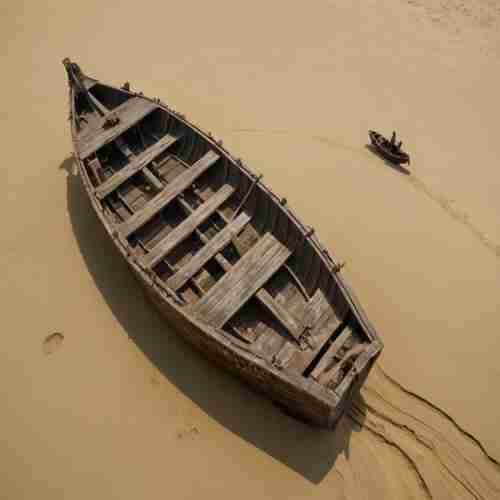}
    \end{minipage}
    \begin{minipage}{0.25\columnwidth}
        \centering
        \includegraphics[width=\linewidth]{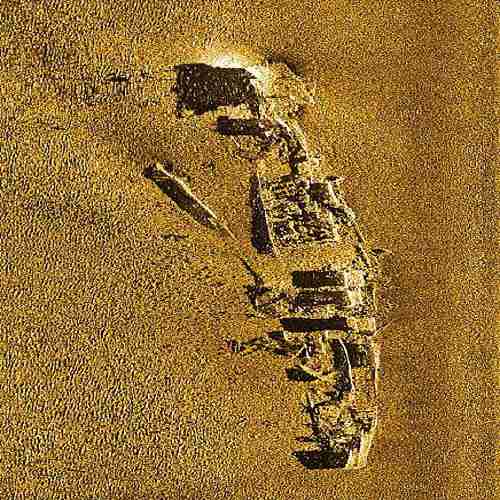}
    \end{minipage}
    \begin{minipage}{0.25\columnwidth}
        \centering
        \includegraphics[width=\linewidth]{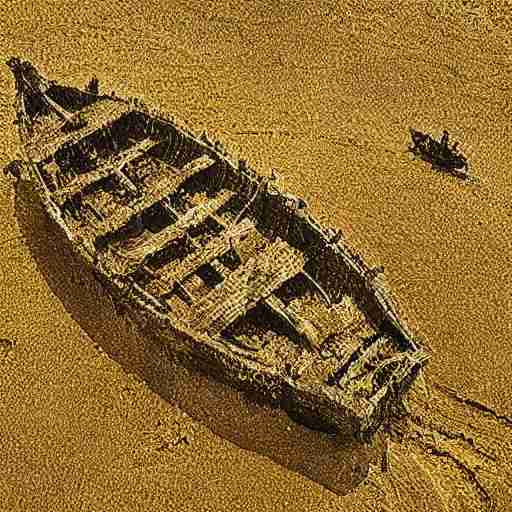}
    \end{minipage}
    
    \begin{minipage}{0.25\columnwidth}
        \centering
        \includegraphics[width=\linewidth]{Images/Supplementary_Style_Injection_Outputs/content/ship_28.jpg}
    \end{minipage}
    \begin{minipage}{0.25\columnwidth}
        \centering
        \includegraphics[width=\linewidth]{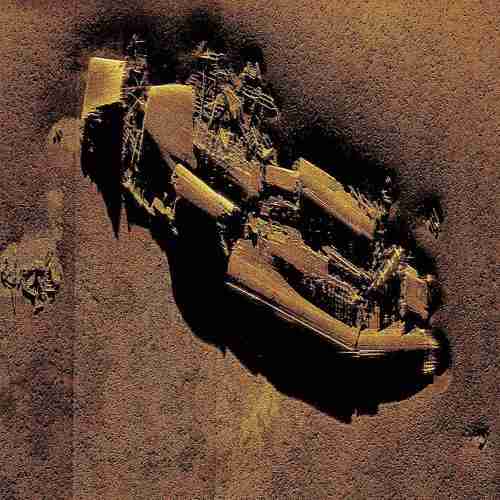}
    \end{minipage}
    \begin{minipage}{0.25\columnwidth}
        \centering
        \includegraphics[width=\linewidth]{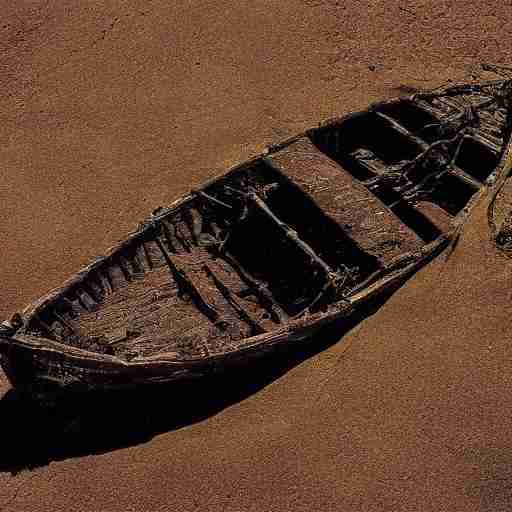}
    \end{minipage}

    \begin{minipage}{0.25\columnwidth}
        \centering
        \includegraphics[width=\linewidth]{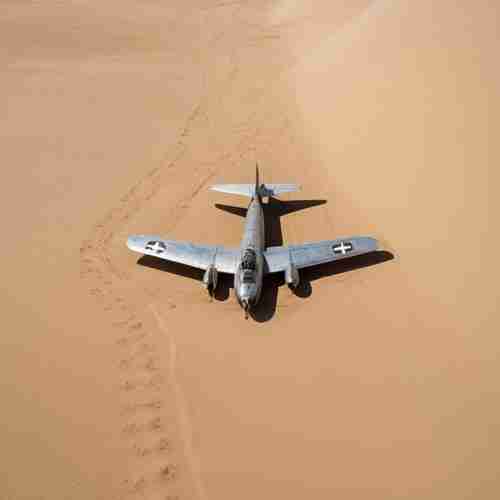}
    \end{minipage}
    \begin{minipage}{0.25\columnwidth}
        \centering
        \includegraphics[width=\linewidth]{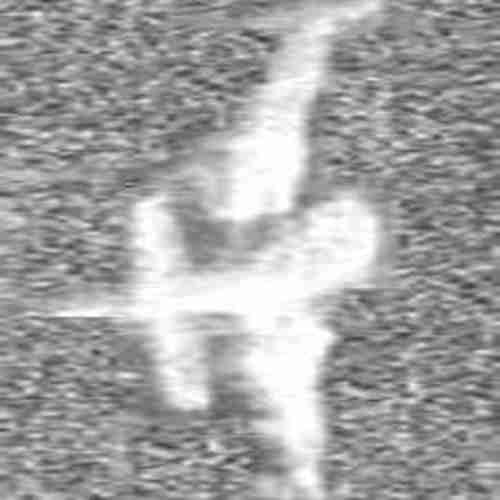}
    \end{minipage}
    \begin{minipage}{0.25\columnwidth}
        \centering
        \includegraphics[width=\linewidth]{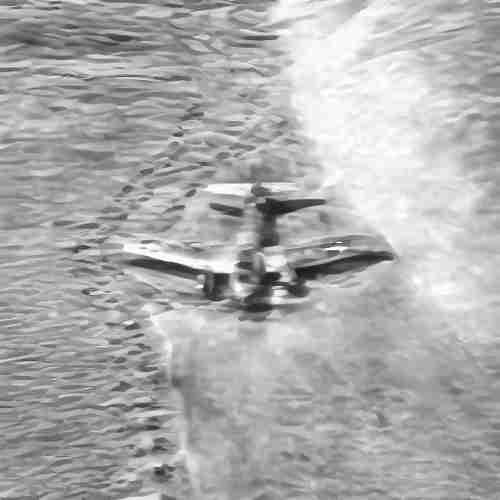}
    \end{minipage}
    
    \begin{minipage}{0.25\columnwidth}
        \centering
        \includegraphics[width=\linewidth]{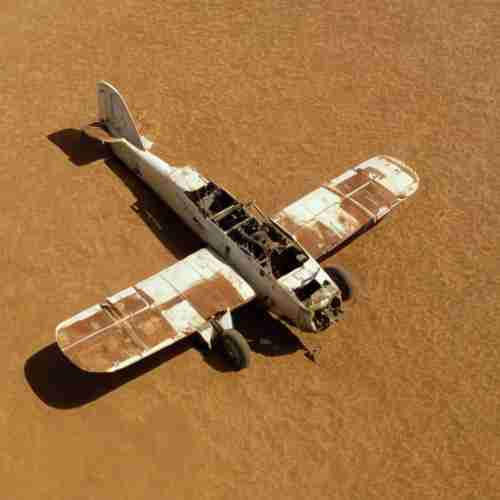}
    \end{minipage}
    \begin{minipage}{0.25\columnwidth}
        \centering
        \includegraphics[width=\linewidth]{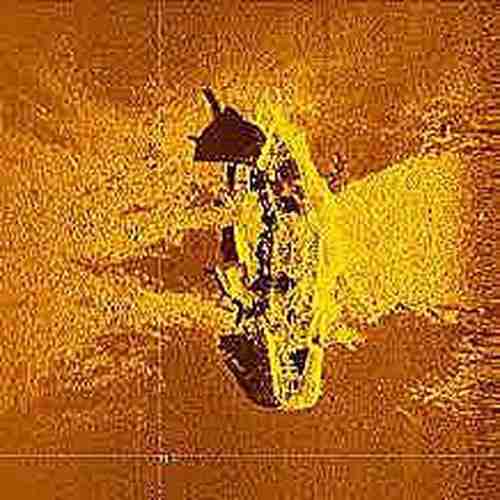}
    \end{minipage}
    \begin{minipage}{0.25\columnwidth}
        \centering
        \includegraphics[width=\linewidth]{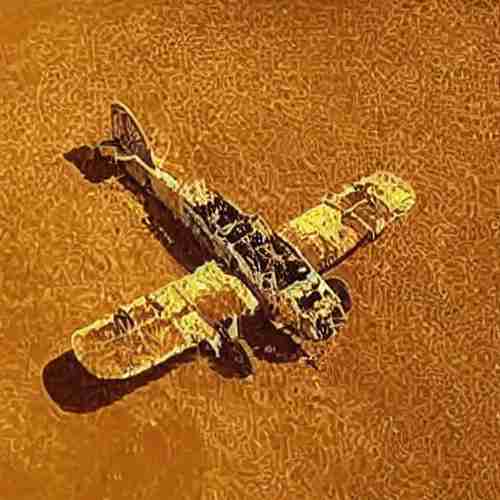}
    \end{minipage}

    % Fourth row of images
    \begin{minipage}{0.25\columnwidth}
        \centering
        \includegraphics[width=\linewidth]{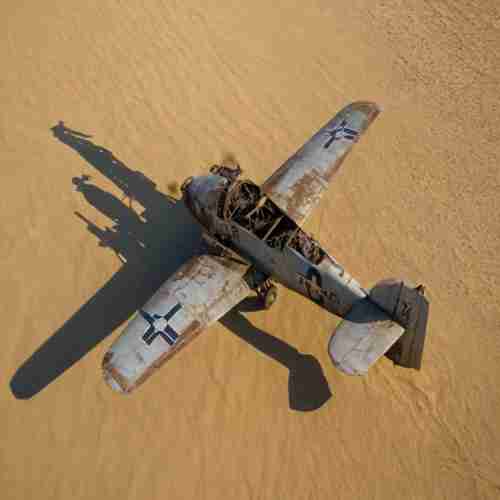}
    \end{minipage}
    \begin{minipage}{0.25\columnwidth}
        \centering
        \includegraphics[width=\linewidth]{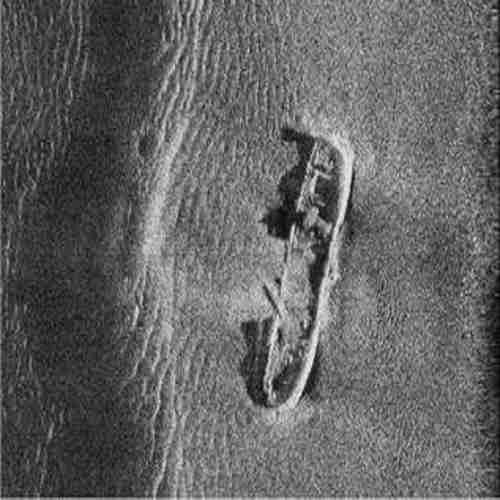}
    \end{minipage}
    \begin{minipage}{0.25\columnwidth}
        \centering
        \includegraphics[width=\linewidth]{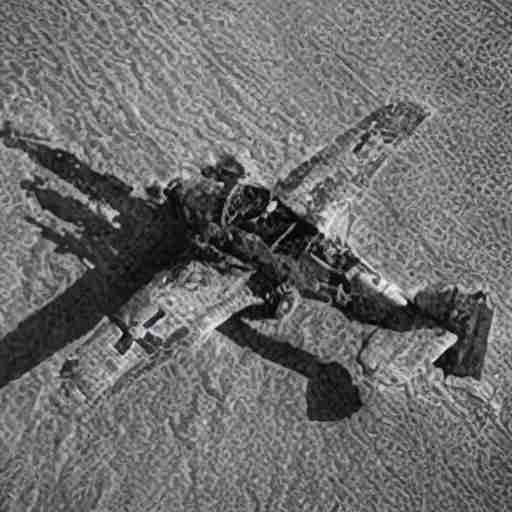}
    \end{minipage}
    
    \vspace{.1cm}
    \caption{The Contents, Styles, and Stylized sonar Images (Outputs) in the style injection process}
    \label{fig: stylized_images}
\end{figure}

The images generated using the style injection are depicted in Fig.~\ref{fig: stylized_images}. The content image is generated from the prompt developed using GPT, the actual image is from a real sonar dataset, and the output is the stylized image from the proposed framework with the value of \(\gamma\) set to 0.5. The stylized output images retain the core structural elements necessary for accurate sonar image interpretation, such as the clear depiction of object boundaries and the representation of noise patterns that are characteristic of underwater acoustic imaging. This ensures that the model does not over-stylize the content, but instead enhances the aesthetic appeal while preserving the scientific accuracy of the generated outputs. Moreover, 
the generated images show clearer details and better feature separation, indicating that style injection with an attention mechanism helps the diffusion model distinguish foreground objects from background noise. This makes the images more similar to real sonar data while adding subtle artistic variations that enhance diversity.

\subsubsection{Sonar Image Classification Model}
\label{subsubsec: sonar image classification model} 
To support the quantitative measure of sonar image synthesis, a classification task is carried out using the generated synthetic data. In particular, we developed classification models leveraging transfer learning on backbone models such as VGG16, ResNet50, DenseNet121, MobileNetV2, Xception, and InceptionResNetV2 With the real, synthetic, and real + synthetic dataset and the performance of the developed image classification model is assessed with real sonar dataset. Refer to Table~\ref{tab: classification models output.} for the test accuracies of various models on real dataset. The model developed with fully synthetic dataset gave a maximum accuracy of 79\%, using the MobileNetV2 as base model. When using only real data, the DenseNet121 model reached the highest accuracy of 96\%, and when using the real + synthetic dataset, DenseNet121 again achieved the best performance with 97\%. This demonstrates that our approach can generate high-quality synthetic sonar data, and the inclusion of synthetic data slightly improves model performance on real datasets.

\begin{table}[H]
    \centering
    \footnotesize
    {\fontsize{8}{10}\selectfont
    \centering
    \begin{tabular}{|c|c|c|c|}
        \hline
        \textbf{Backbone Model} & \textbf{Real} & \textbf{Synthetic} & \textbf{Real + Synthetic} \\
        \hline
        VGG16 & 95\% & 58\% & 94\% \\
        ResNet50 & 90\% & 29\% & 82\% \\
        DenseNet121 & \textbf{96\%} & 75\% & \textbf{97\%} \\
        MobileNetV2 & 92\% & \textbf{79\%} & 95\% \\
        Xception & 95\% & 68\% & 95\% \\
        InceptionResNetV2 & 95\% & 48\% & 94\% \\
        \hline 
    \end{tabular}
    } %end footnotesize
    \vspace{.1cm}
    \centering
    \caption{Test Accuracy on Real Dataset}
    \label{tab: classification models output.}
\end{table}

\subsection{Phase-2 \& 3: Fine-Tuning Results}
\label{subsec: fine-tunning results}
Training or fine-tuning stable diffusion models typically demands extensive computational resources and large amounts of annotated data. However, due to computational constraints and limited access to experts for annotating generated images, we evaluated our framework as a pilot study, utilizing a dataset of 30 images per label. In total, the dataset consists of six different objects with 18 object variations across six seafloor environments used for model training.

\begin{figure}[H]
    \vspace{-0.2cm}
    \centering
    \begin{minipage}{0.22\columnwidth}
        \centering
        \fontsize{8}{10}\selectfont\text{Seafloor} % Adjust the size here
    \end{minipage}
    \begin{minipage}{0.22\columnwidth}
        \centering
        \fontsize{8}{10}\selectfont\text{Mine} % Adjust the size here
    \end{minipage}
    \begin{minipage}{0.22\columnwidth}
        \centering
        \fontsize{8}{10}\selectfont\text{Plane} % Adjust the size here
    \end{minipage}
    \begin{minipage}{0.22\columnwidth}
        \centering
        \fontsize{8}{10}\selectfont\text{Ship} % Adjust the size here
    \end{minipage}
    \vspace{0.05cm}
    \begin{minipage}{0.22\columnwidth}
        \centering
        \includegraphics[width=\linewidth]{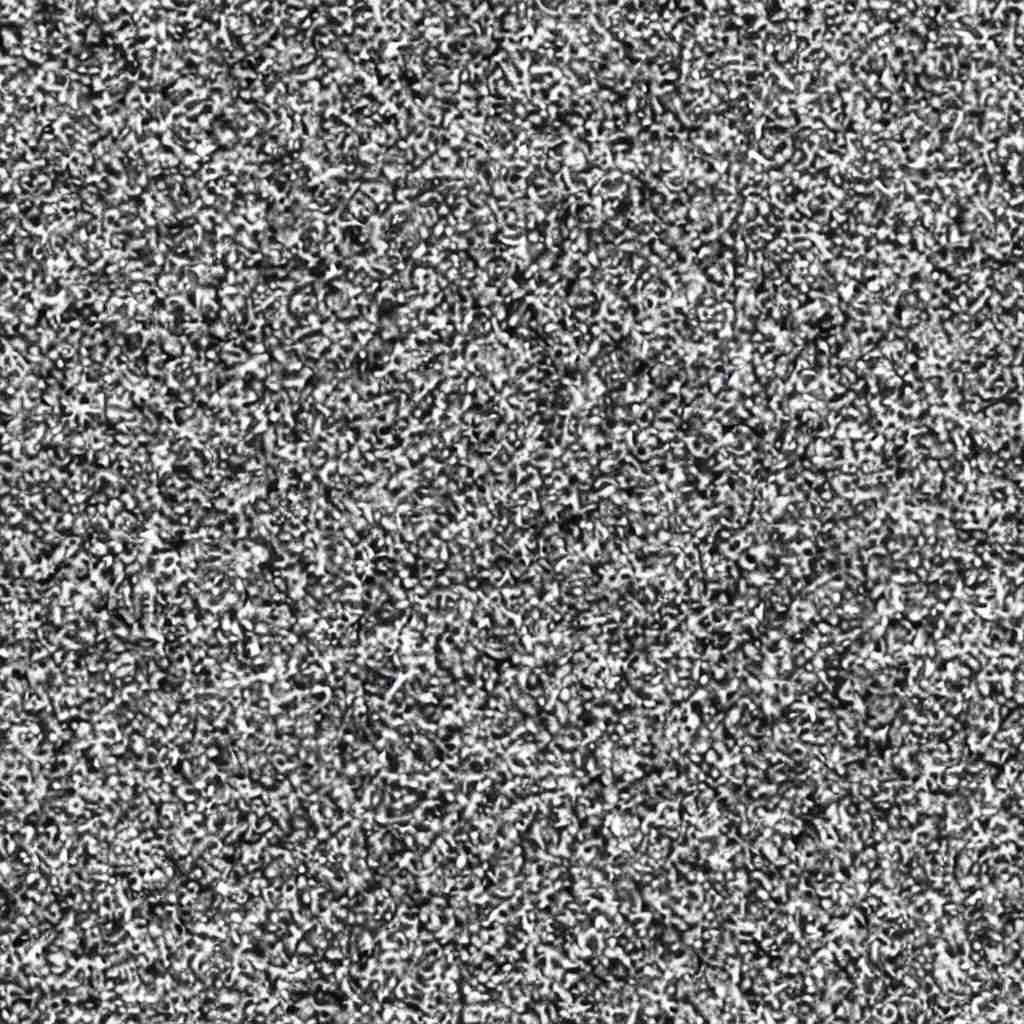}
    \end{minipage}
    %\hspace{0.0001\columnwidth}
    \begin{minipage}{0.22\columnwidth}
        \centering
        \includegraphics[width=\linewidth]{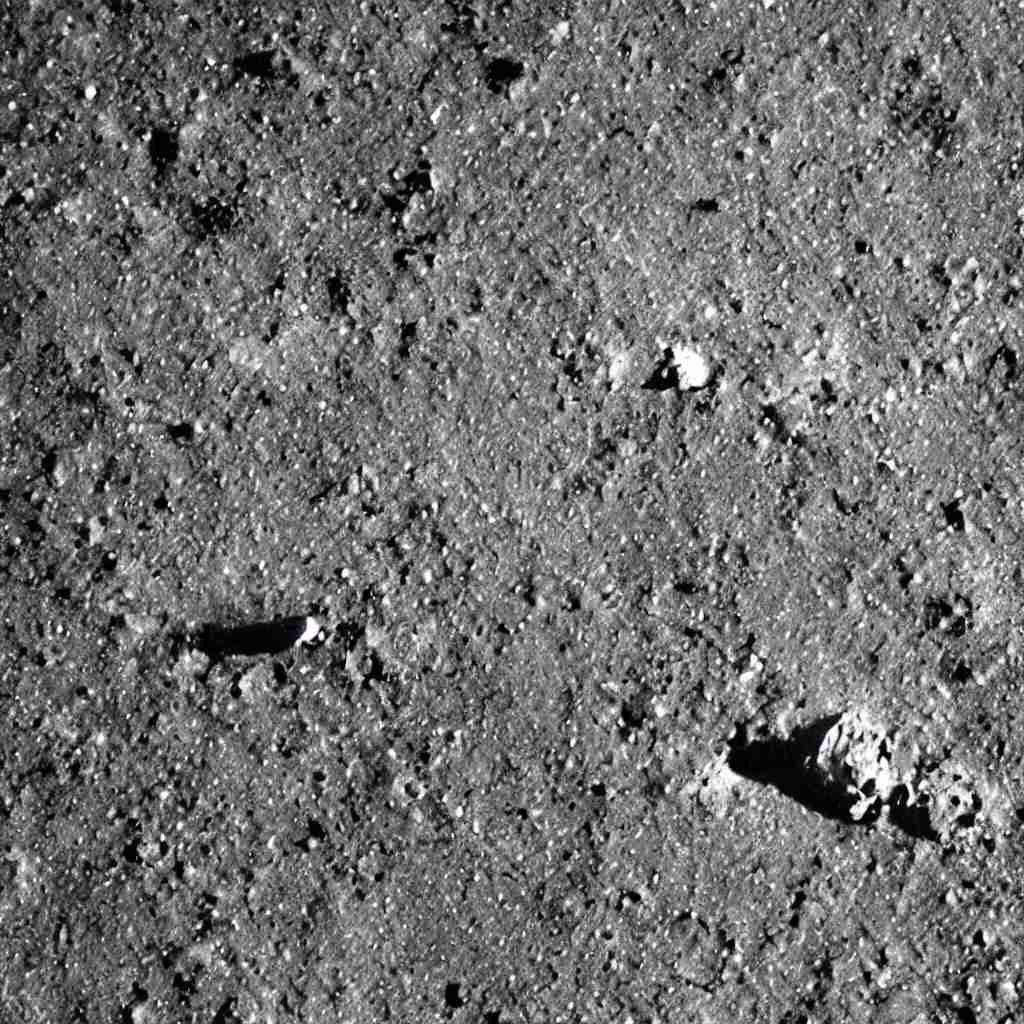}
    \end{minipage}
    %\hspace{0.0001\columnwidth}
    \begin{minipage}{0.22\columnwidth}
        \centering
        \includegraphics[width=\linewidth]{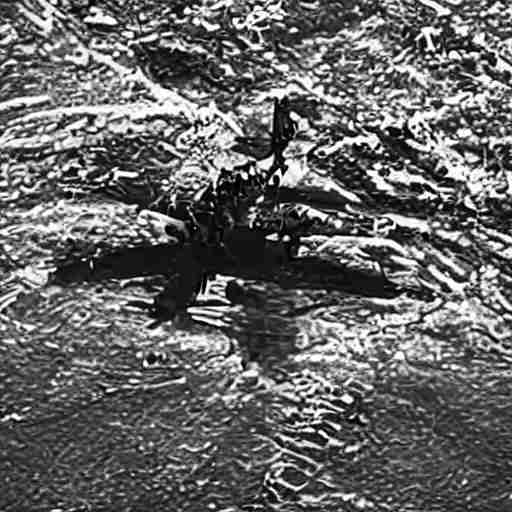}
    \end{minipage}
    %\hspace{0.001\columnwidth}
    \begin{minipage}{0.22\columnwidth}
        \centering
        \includegraphics[width=\linewidth]{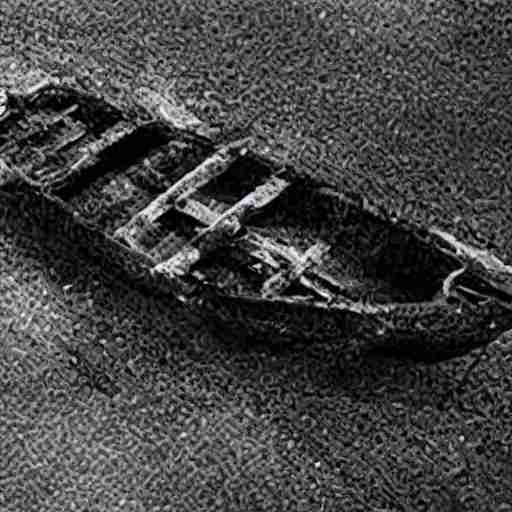}
    \end{minipage}
    \vspace{.1cm}
    \caption{Generated Samples from the fine-tuned model}
    \label{fig: generated samples from the fine-tuned model}
\end{figure}

\begin{figure*}[h!]
    \vspace{-0.3cm}
    \centering
    \begin{minipage}{0.095\textwidth} % Slightly smaller to fit all images
        \centering
        \fontsize{8}{10}\selectfont\text{500}
    \end{minipage}
    \begin{minipage}{0.095\textwidth}
        \centering
        \fontsize{8}{10}\selectfont\text{1000}
    \end{minipage}
    \begin{minipage}{0.095\textwidth}
        \centering
        \fontsize{8}{10}\selectfont\text{1500}
    \end{minipage}
    \begin{minipage}{0.095\textwidth}
        \centering
        \fontsize{8}{10}\selectfont\text{3000}
    \end{minipage}
    \begin{minipage}{0.095\textwidth}
        \centering
        \fontsize{8}{10}\selectfont\text{5000}
    \end{minipage}
    \begin{minipage}{0.095\textwidth}
        \centering
        \fontsize{8}{10}\selectfont\text{6500}
    \end{minipage}
    \begin{minipage}{0.095\textwidth}
        \centering
        \fontsize{8}{10}\selectfont\text{7500}
    \end{minipage}
    \begin{minipage}{0.095\textwidth}
        \centering
        \fontsize{8}{10}\selectfont\text{10000}
    \end{minipage}
    \begin{minipage}{0.095\textwidth}
        \centering
        \fontsize{8}{10}\selectfont\text{12500}
    \end{minipage}
    \begin{minipage}{0.095\textwidth}
        \centering
        \fontsize{8}{10}\selectfont\text{15000}
    \end{minipage}
    \vspace{0.05cm}
    \centering
    \begin{minipage}{0.095\textwidth}
        \centering
        \includegraphics[width=\linewidth]{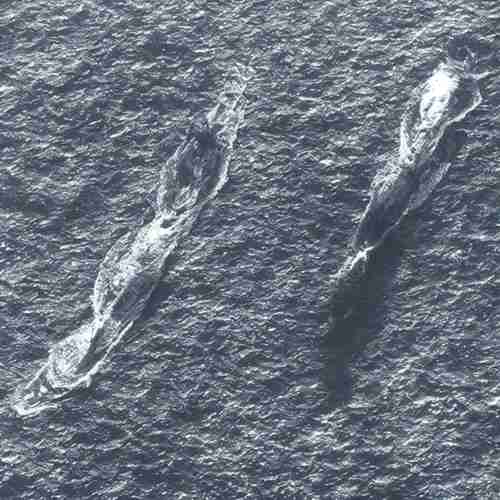}
    \end{minipage}
    \begin{minipage}{0.095\textwidth}
        \centering
        \includegraphics[width=\linewidth]{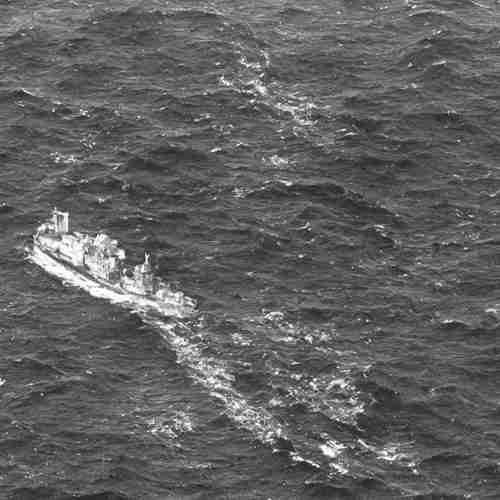}
    \end{minipage}
    \begin{minipage}{0.095\textwidth}
        \centering
        \includegraphics[width=\linewidth]{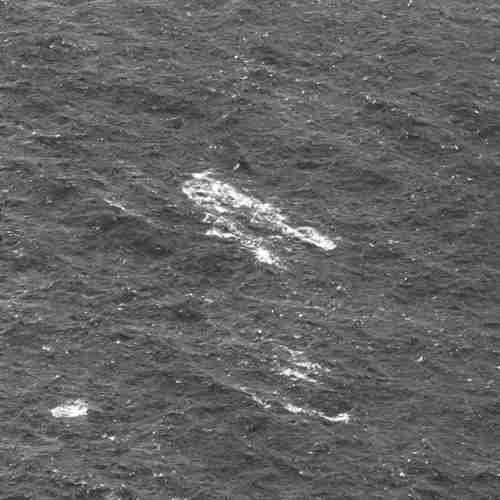}
    \end{minipage}
    \begin{minipage}{0.095\textwidth}
        \centering
        \includegraphics[width=\linewidth]{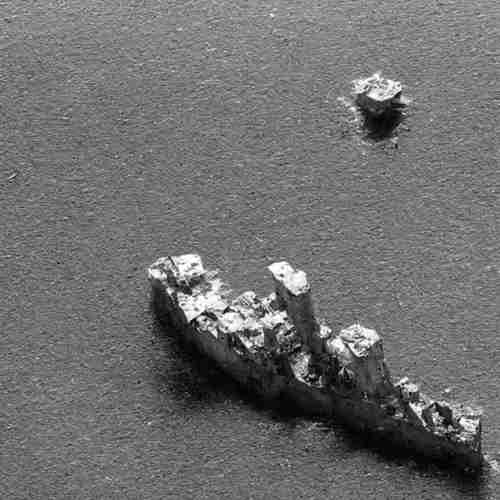}
    \end{minipage}
    \begin{minipage}{0.095\textwidth}
        \centering
        \includegraphics[width=\linewidth]{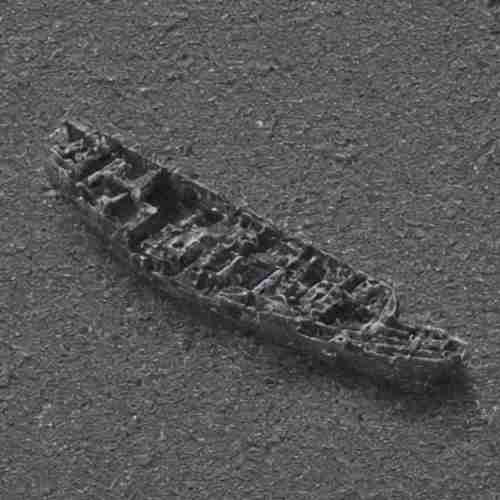}
    \end{minipage}
    \begin{minipage}{0.095\textwidth}
        \centering
        \includegraphics[width=\linewidth]{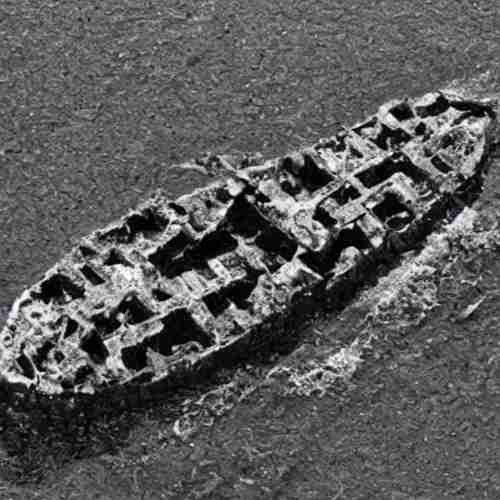}
    \end{minipage}
    \begin{minipage}{0.095\textwidth}
        \centering
        \includegraphics[width=\linewidth]{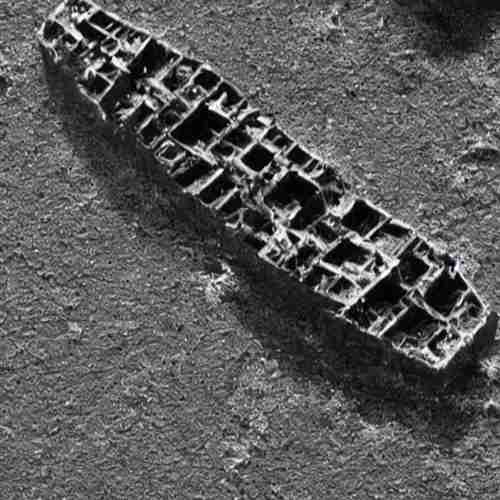}
    \end{minipage}
    \begin{minipage}{0.095\textwidth}
        \centering
        \includegraphics[width=\linewidth]{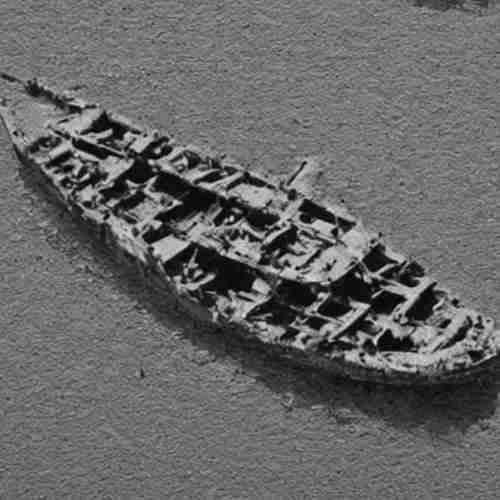}
    \end{minipage}
    \begin{minipage}{0.095\textwidth}
        \centering
        \includegraphics[width=\linewidth]{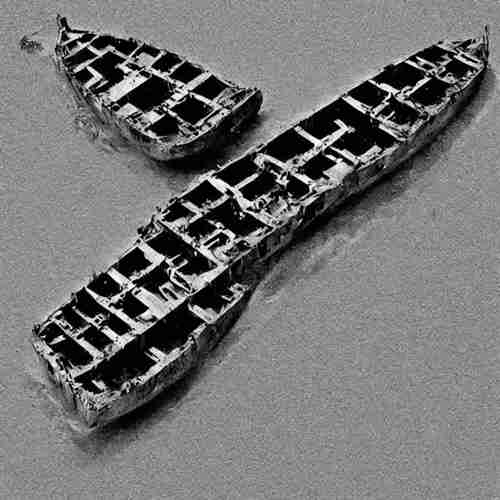}
    \end{minipage}
    \begin{minipage}{0.095\textwidth}
        \centering
        \includegraphics[width=\linewidth]{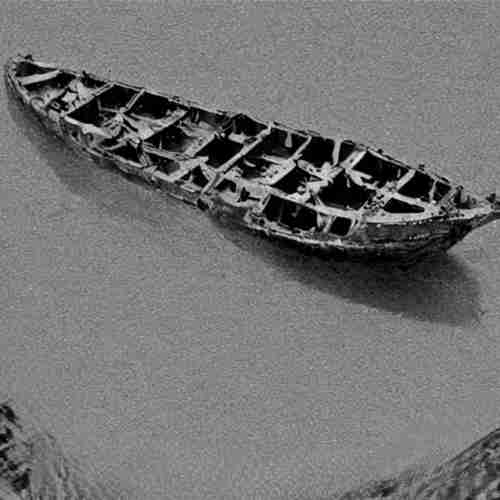}
    \end{minipage}

    \vspace{0.05cm}

    \centering
    \begin{minipage}{0.095\textwidth}
        \centering
        \includegraphics[width=\linewidth]{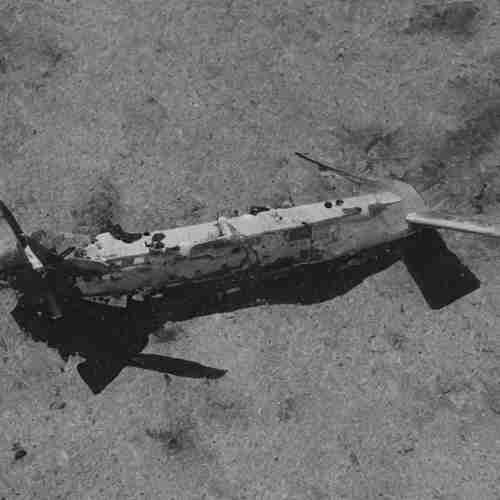}
    \end{minipage}
    \begin{minipage}{0.095\textwidth}
        \centering
        \includegraphics[width=\linewidth]{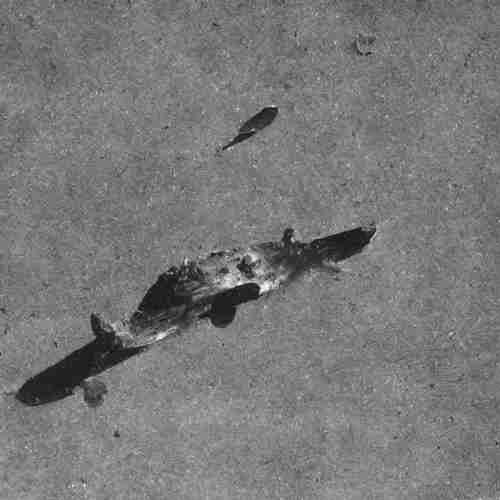}
    \end{minipage}
    \begin{minipage}{0.095\textwidth}
        \centering
        \includegraphics[width=\linewidth]{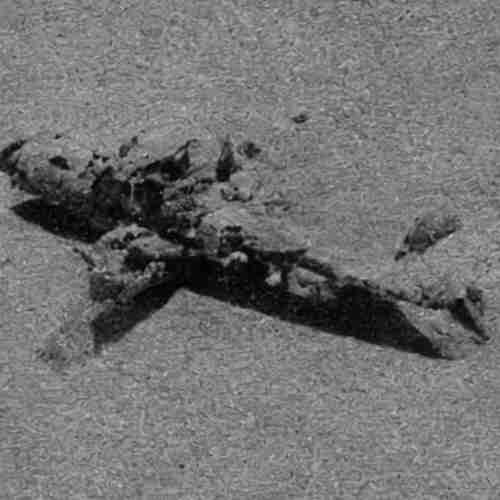}
    \end{minipage}
    \begin{minipage}{0.095\textwidth}
        \centering
        \includegraphics[width=\linewidth]{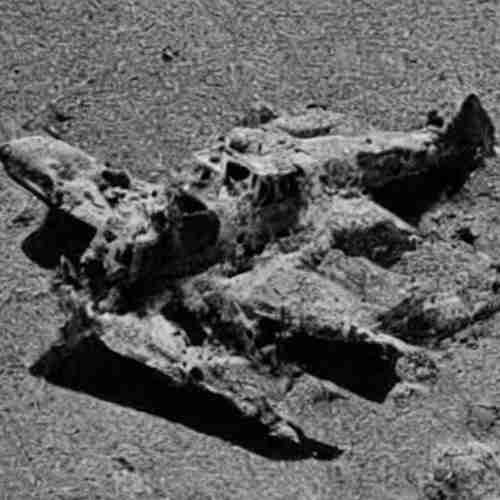}
    \end{minipage}
    \begin{minipage}{0.095\textwidth}
        \centering
        \includegraphics[width=\linewidth]{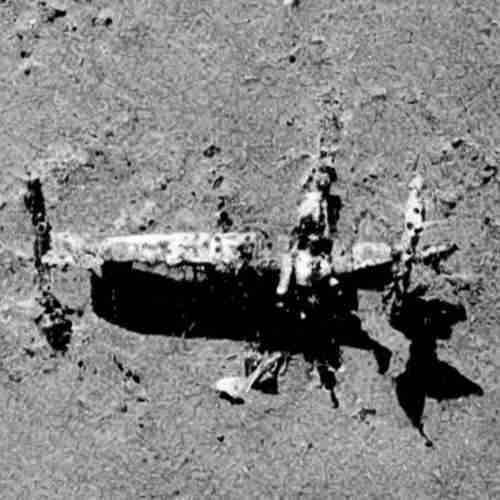}
    \end{minipage}
    \begin{minipage}{0.095\textwidth}
        \centering
        \includegraphics[width=\linewidth]{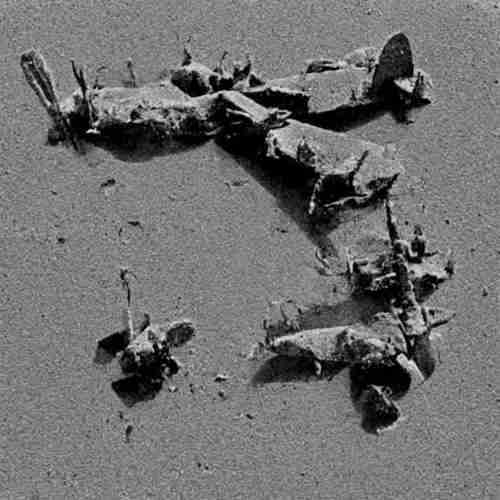}
    \end{minipage}
    \begin{minipage}{0.095\textwidth}
        \centering
        \includegraphics[width=\linewidth]{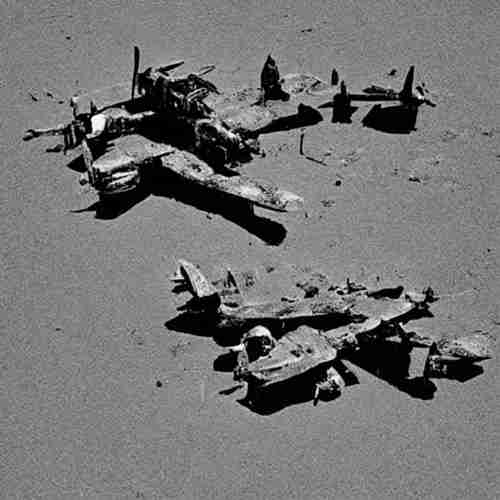}
    \end{minipage}
    \begin{minipage}{0.095\textwidth}
        \centering
        \includegraphics[width=\linewidth]{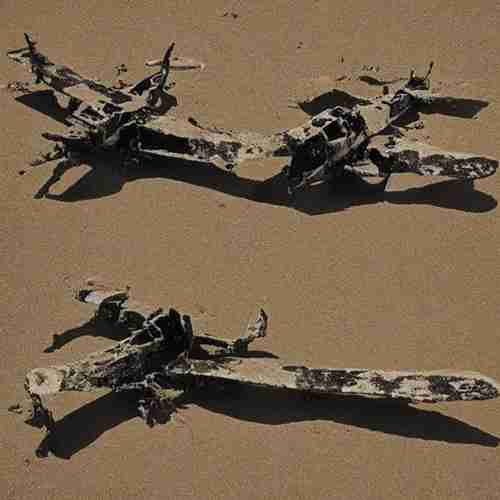}
    \end{minipage}
    \begin{minipage}{0.095\textwidth}
        \centering
        \includegraphics[width=\linewidth]{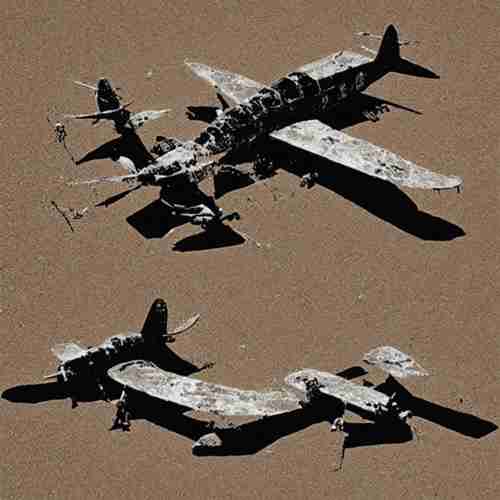}
    \end{minipage}
    \begin{minipage}{0.095\textwidth}
        \centering
        \includegraphics[width=\linewidth]{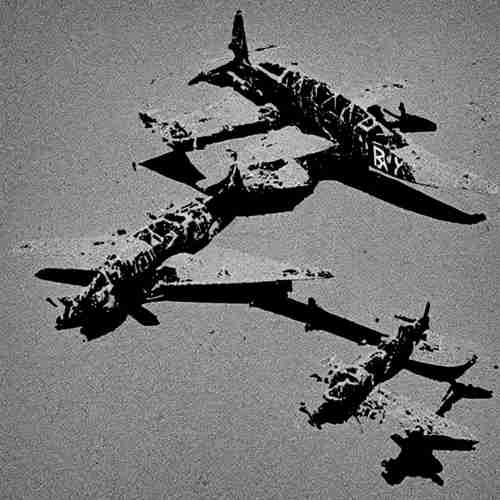}
    \end{minipage}

    \vspace{.1cm}
    \caption{Generated Images w.r.t training steps (Row-1: Number of Training steps, Row-2: Ship generated using the prompt `image of SH34* ship on AP238* seabed', Row-3: Plane generated using the prompt `Image of a PL71* plane in a sandy AS25* seabed, with a broken tail and shadow extending to the front and left')}
    \label{fig: training steps}
\end{figure*}

We qualitatively evaluated the performance of our model, with the generated images shown in Fig.~\ref{fig: generated samples from the fine-tuned model} and Fig.~\ref{fig: training steps}. The model is trained using a dual diffusion framework, comprising coarse and fine image generation phases. In the coarse phase (Phase-2), the model produces images, outlining the basic structure of the sonar scene using detailed and pre-determined prompts. As the training progresses into the fine phase (Phase-3), the model generates higher-resolution images using generic prompts and even multiple objects where applicable.

The proposed hierarchical framework incorporates a Visual Language Model (VLM) to generate descriptions that enhance the image generation process. As detailed in Table~\ref{table:descriptions_from_the_llm}, the VLM supports GPT to provide both low-level descriptions (e.g., shape, size, texture of objects) and high-level descriptions (e.g., object relationships and interactions with the environment) to refine the stable diffusion model.

\subsection{Ablation study}
\label{sec: ablation study}
We conducted an in-depth analysis to regulate the results generated by the trained model. The style-injection ablation study is extensively discussed in Section \ref{subsec: style-injection ablation study}, while the fine-tuning ablation study is detailed in Section \ref{subsec: fine-tuning ablation study}.

\subsubsection{Phase-1: Style-Injection (Image-to-Image)}

\label{subsec: style-injection ablation study}
\begin{figure}[H]
    \vspace{-0.2cm}
    \centering
    \begin{minipage}{0.22\columnwidth}
        \centering
        \fontsize{8}{10}\selectfont\text{$\gamma=0.9$} % Adjust the size here
    \end{minipage}
    \begin{minipage}{0.22\columnwidth}
        \centering
        \fontsize{8}{10}\selectfont\text{$\gamma=0.75$} % Adjust the size here
    \end{minipage}
    \begin{minipage}{0.22\columnwidth}
        \centering
        \fontsize{8}{10}\selectfont\text{$\gamma=0.5$} % Adjust the size here
    \end{minipage}
    \begin{minipage}{0.22\columnwidth}
        \centering
        \fontsize{8}{10}\selectfont\text{$\gamma=0.3$} % Adjust the size here
    \end{minipage}
    \vspace{0.05cm}
    \centering
    \begin{minipage}{0.2\columnwidth}
        \centering
        \includegraphics[width=\linewidth]{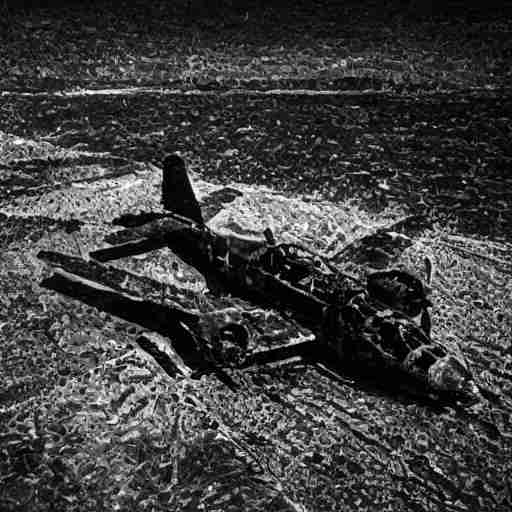}
    \end{minipage}
    \begin{minipage}{0.2\columnwidth}
        \centering
        \includegraphics[width=\linewidth]{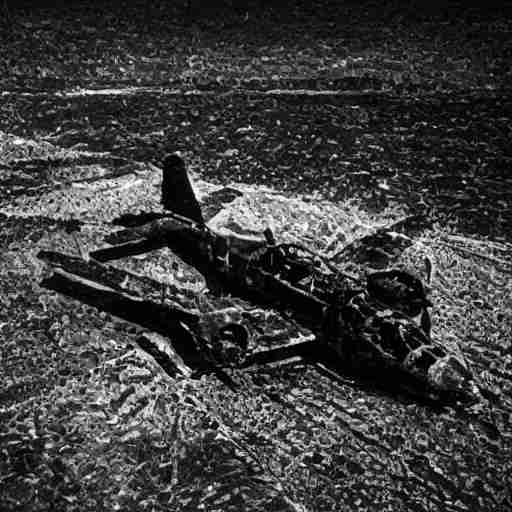}
    \end{minipage}
    \begin{minipage}{0.2\columnwidth}
        \centering
        \includegraphics[width=\linewidth]{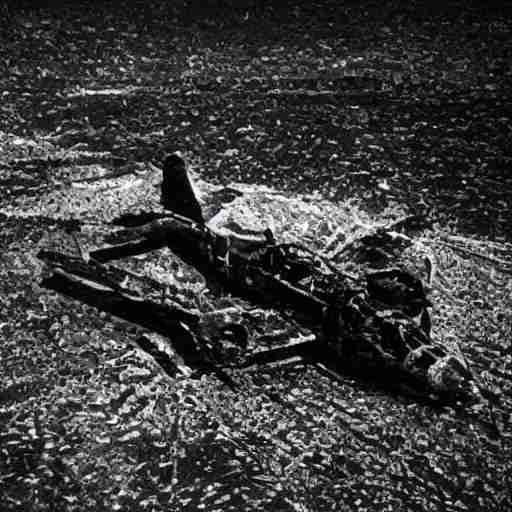}
    \end{minipage}
    \begin{minipage}{0.2\columnwidth}
        \centering
        \includegraphics[width=\linewidth]{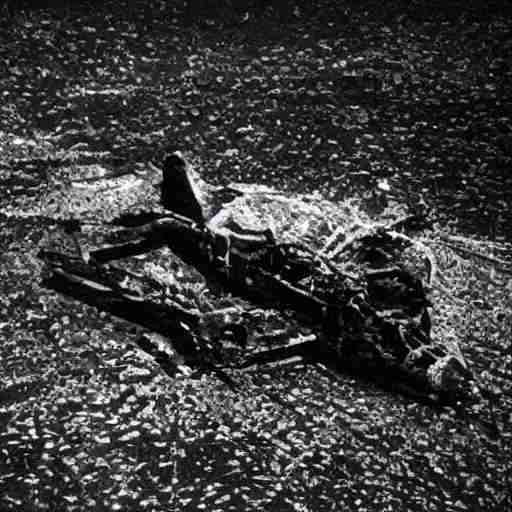}
    \end{minipage}
    \vspace{0.05cm}
    \centering
    \begin{minipage}{0.2\columnwidth}
        \centering
        \includegraphics[width=\linewidth]{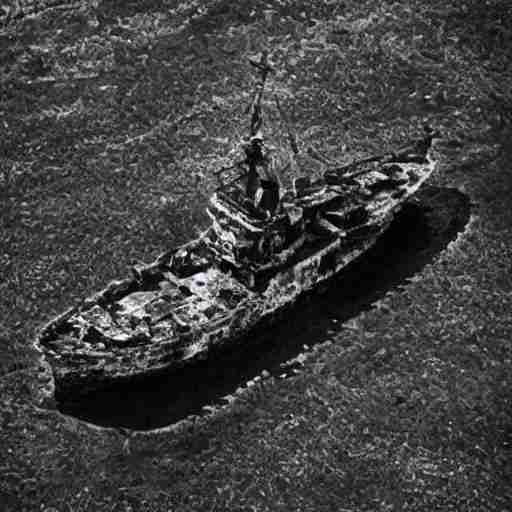}
    \end{minipage}
    \begin{minipage}{0.2\columnwidth}
        \centering
        \includegraphics[width=\linewidth]{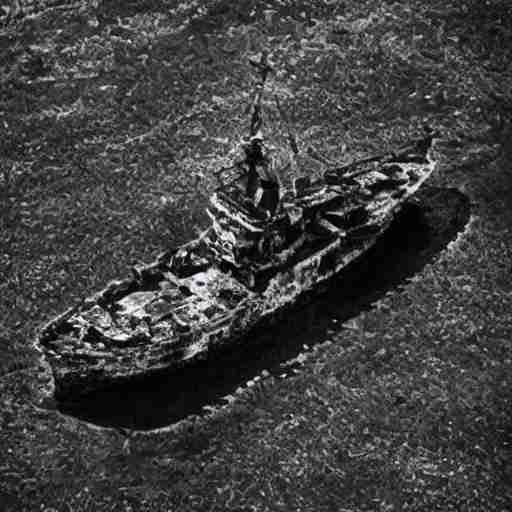}
    \end{minipage}
    \begin{minipage}{0.2\columnwidth}
        \centering
        \includegraphics[width=\linewidth]{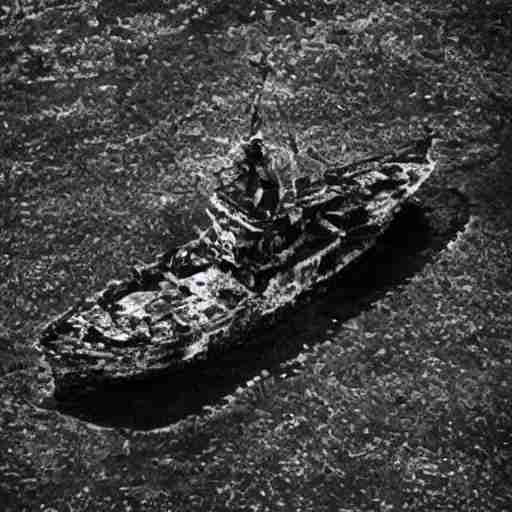}
    \end{minipage}
    \begin{minipage}{0.2\columnwidth}
        \centering
        \includegraphics[width=\linewidth]{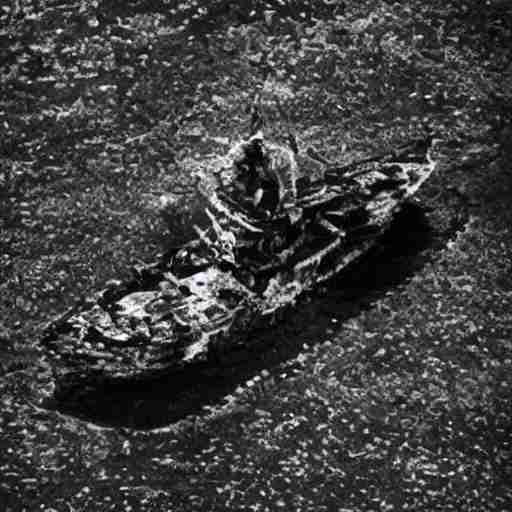}
    \end{minipage}
    \vspace{.1cm}
    \caption{Style and Content Fidelity ($\gamma$); Lower values improve style fidelity but may reduce content fidelity.}
    \label{fig:style_and_content_fidelity}
\end{figure}

The blending ratio of content and style is controlled by the parameter \(\gamma\). We conducted a comparative study to examine how it impacts the quality of the image produced through style injection, as shown in Fig.~\ref{fig:style_and_content_fidelity}. 

The generated images appear to be similar to each other. However, qualitative metrics such as SSIM and PSNR (refer to Table~\ref{table:ssim and psnr w.r.t alpha}) were computed for images generated with \(\gamma = \{0.9, 0.75, 0.5, 0.3\}\) and found that as \(\gamma\) decreases, there is a noticeable improvement in both SSIM and PSNR, indicating a gradual enhancement in visual quality and fidelity. For our main model, we set \(\gamma = 0.5\) as the default value, striking a balance between preserving content and injecting stylistic information while avoiding excessive distortion or loss of essential content features.

\begin{table}[h!]
    \centering
    \begin{tabular}{|c|c|c|}
        \hline
        \textbf{$\gamma$} & \textbf{SSIM} & \textbf{PSNR} \\
        \hline
        0.9 & 0.2360 & 11.1372 \\
        0.75 & 0.2535 & 11.3949 \\
        0.5 & 0.2696 & 11.5774 \\
        0.3 & 0.2761 & 11.6498 \\
        \hline
    \end{tabular}
    \vspace{.1cm}
    \caption{SSIM \& PSNR w.r.t. fidelity ($\gamma$) for the images in Fig.~\ref{fig:style_and_content_fidelity}}
    \label{table:ssim and psnr w.r.t alpha}
\end{table} 

\vspace{-0.5cm}
\subsubsection{Phase-2 \& 3: Fine-Tuning (Text-to--Image)}
\label{subsec: fine-tuning ablation study}

The number of training steps significantly impacts the quality of the generated images. To verify the same, a qualitative study on the outputs of the trained model (see Fig.~\ref{fig: training steps}) is carried out. It is observed that at lower step counts, the images tend to be coarse, with artifacts or incomplete object representations. Increasing the number of steps leads to sharper images with clearer object boundaries and better textures, allowing for more accurate and realistic depictions of underwater scenes. 
 
For the ship category, the model started generating sonar images only after 3000 training steps. Between 5000 and 10000 training steps, the quality of the generated images was good. However, after 10000 training steps, the model failed to produce the expected output and instead generated multiple ships. In the context of planes, the given prompt ``Image of a PL71* plane in a sandy AS25* seabed, with a broken tail and shadow extending to the front and left" is relatively intricate. The model started generating sonar images after 1500 training iterations and consistently produced satisfactory outcomes until 5000 training iterations. However, after 6500 training iterations, the model yielded unsatisfactory outputs with multiple aircraft.

\begin{figure}[ht!]
    \centering
    % First row of images
    \begin{minipage}{0.22\columnwidth}
        \includegraphics[width=\linewidth]{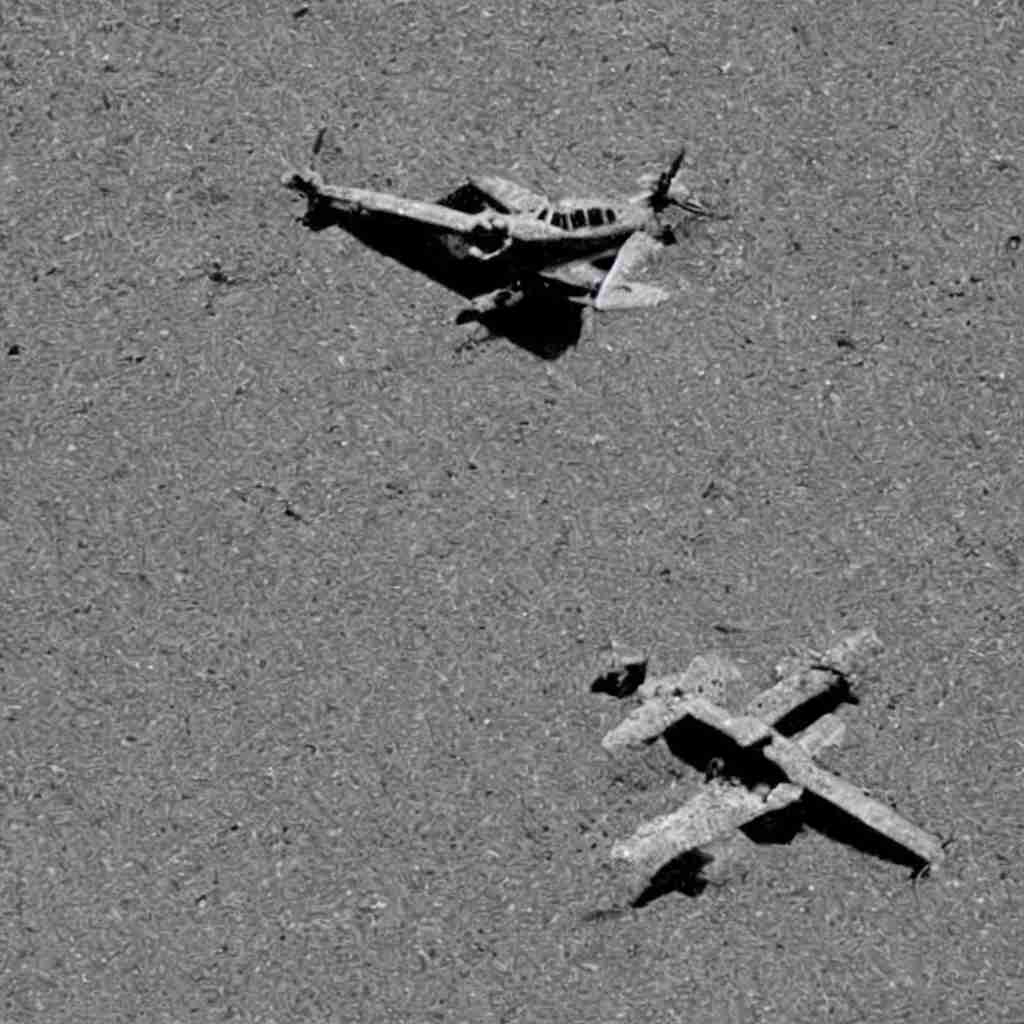}
    \end{minipage}
    \begin{minipage}{0.22\columnwidth}
        \includegraphics[width=\linewidth]{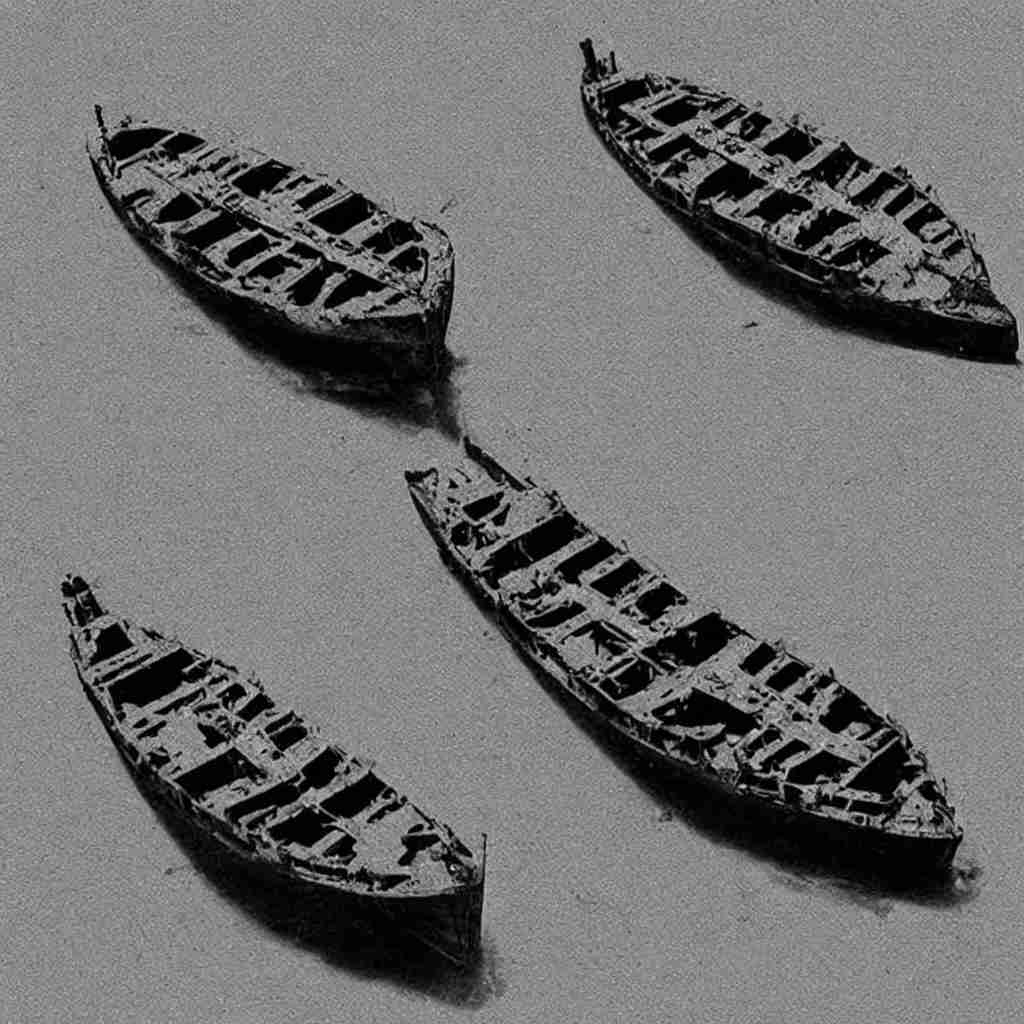}
    \end{minipage}
    \begin{minipage}{0.22\columnwidth}
        \includegraphics[width=\linewidth]{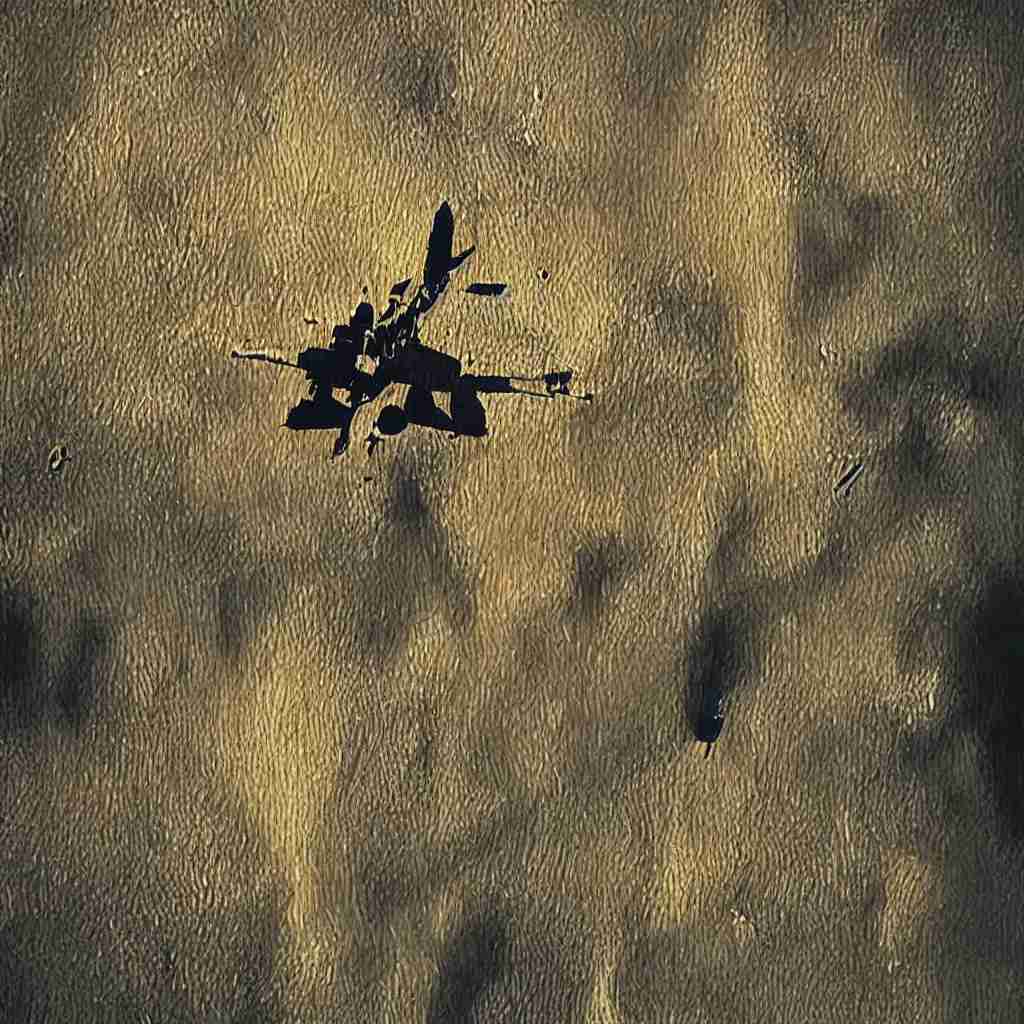}
    \end{minipage}
    \begin{minipage}{0.22\columnwidth}
        \includegraphics[width=\linewidth]{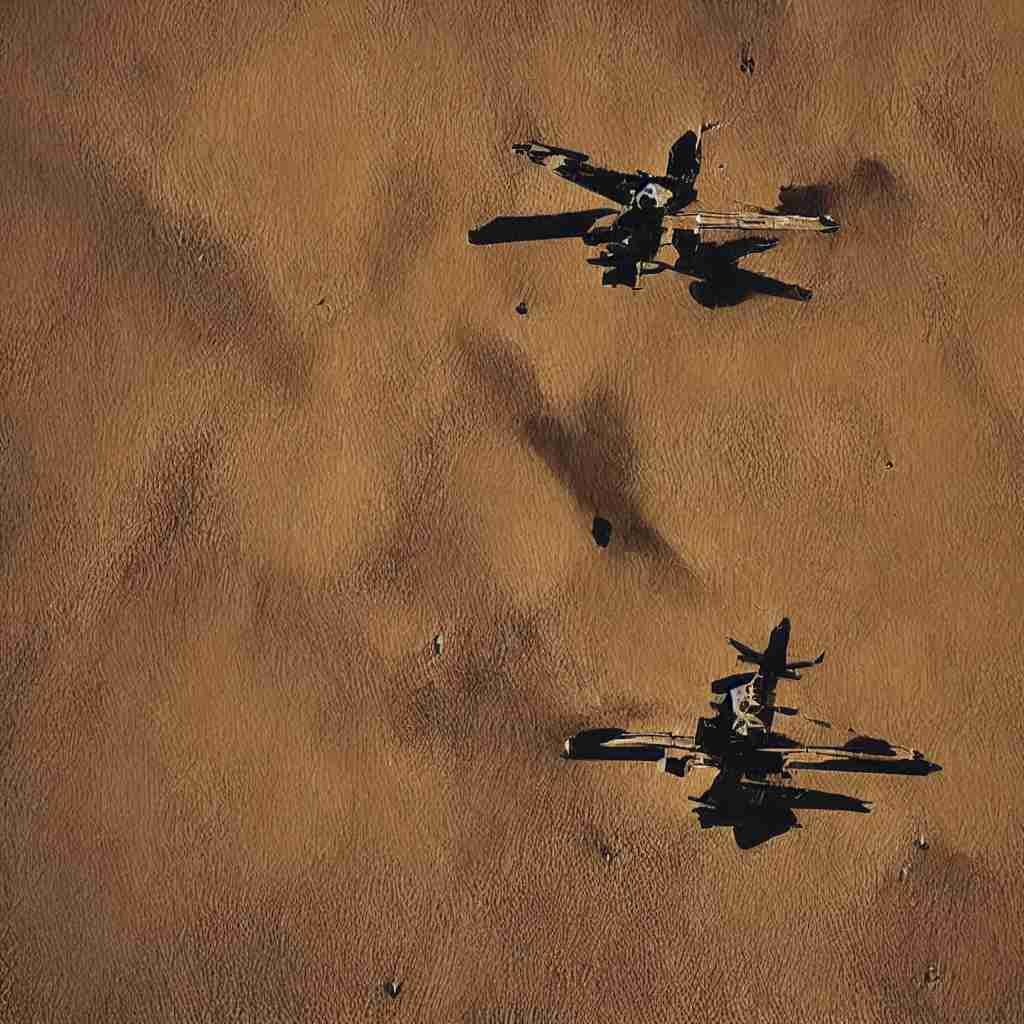}
    \end{minipage}
    \vspace{0.05cm} % Space between rows of images
    
    % Second row of images
    \begin{minipage}{0.22\columnwidth}
        \includegraphics[width=\linewidth]{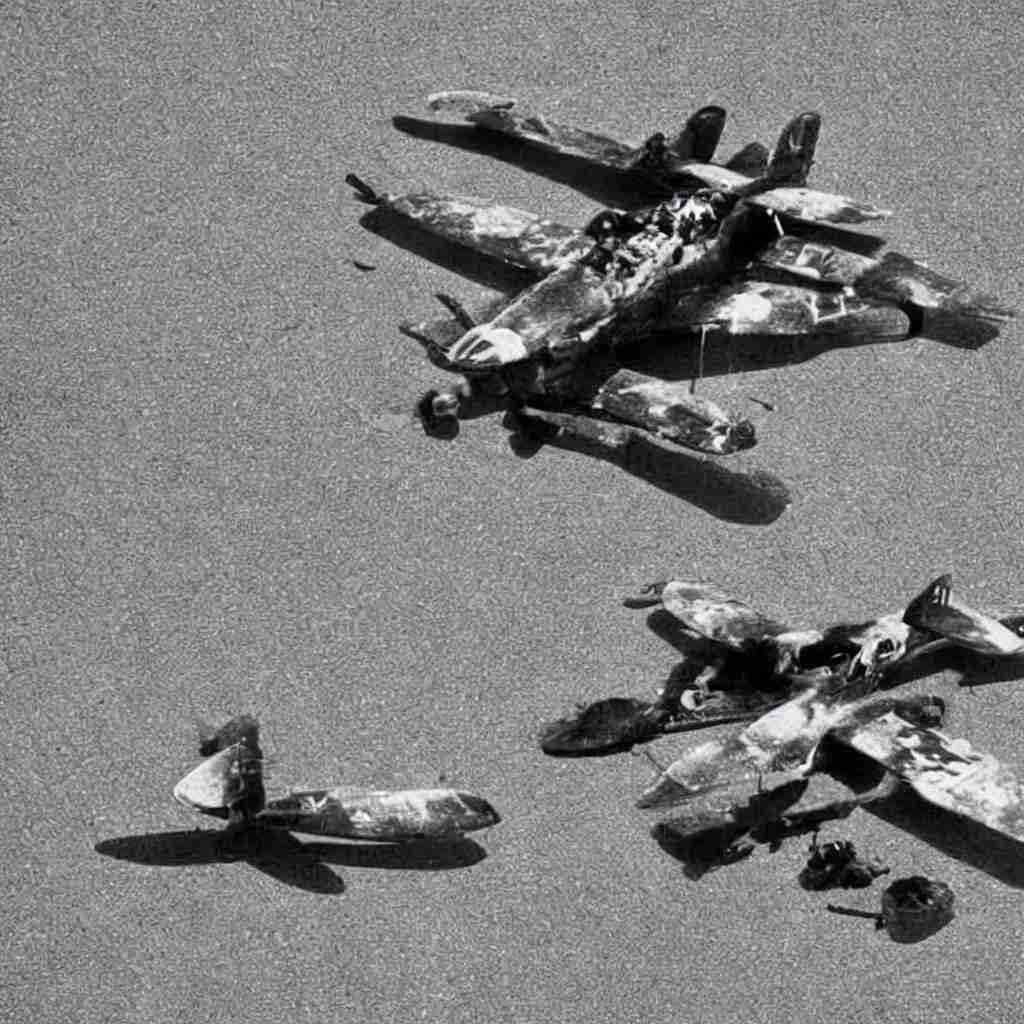}
    \end{minipage}
    \begin{minipage}{0.22\columnwidth}
        \includegraphics[width=\linewidth]{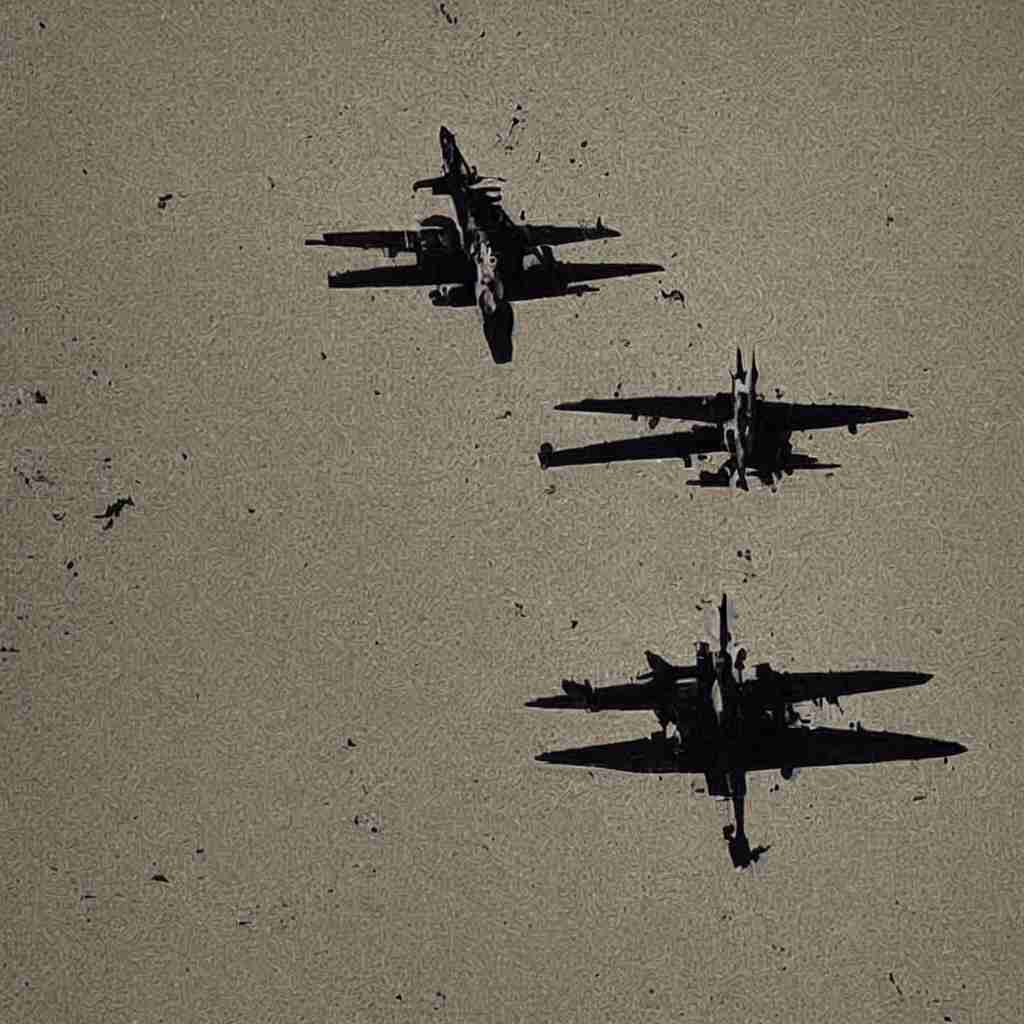}
    \end{minipage}
    \begin{minipage}{0.22\columnwidth}
        \includegraphics[width=\linewidth]{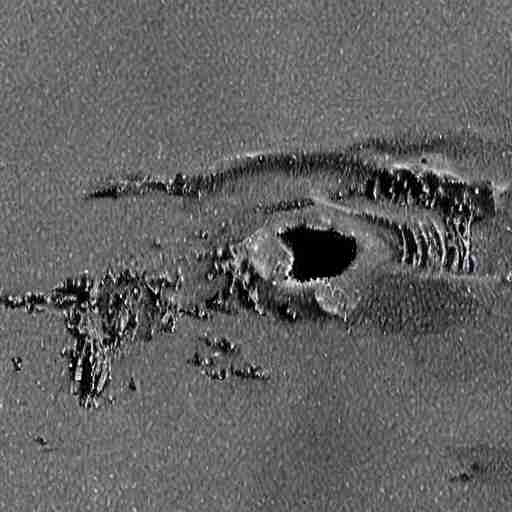}
    \end{minipage}
    \begin{minipage}{0.22\columnwidth}
        \includegraphics[width=\linewidth]{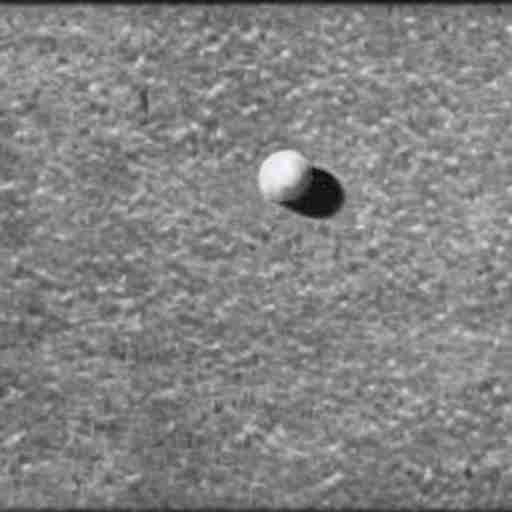}
    \end{minipage}
    \vspace{0.05cm} % Space between rows of images
    
    % Third row of images
    \begin{minipage}{0.22\columnwidth}
        \includegraphics[width=\linewidth]{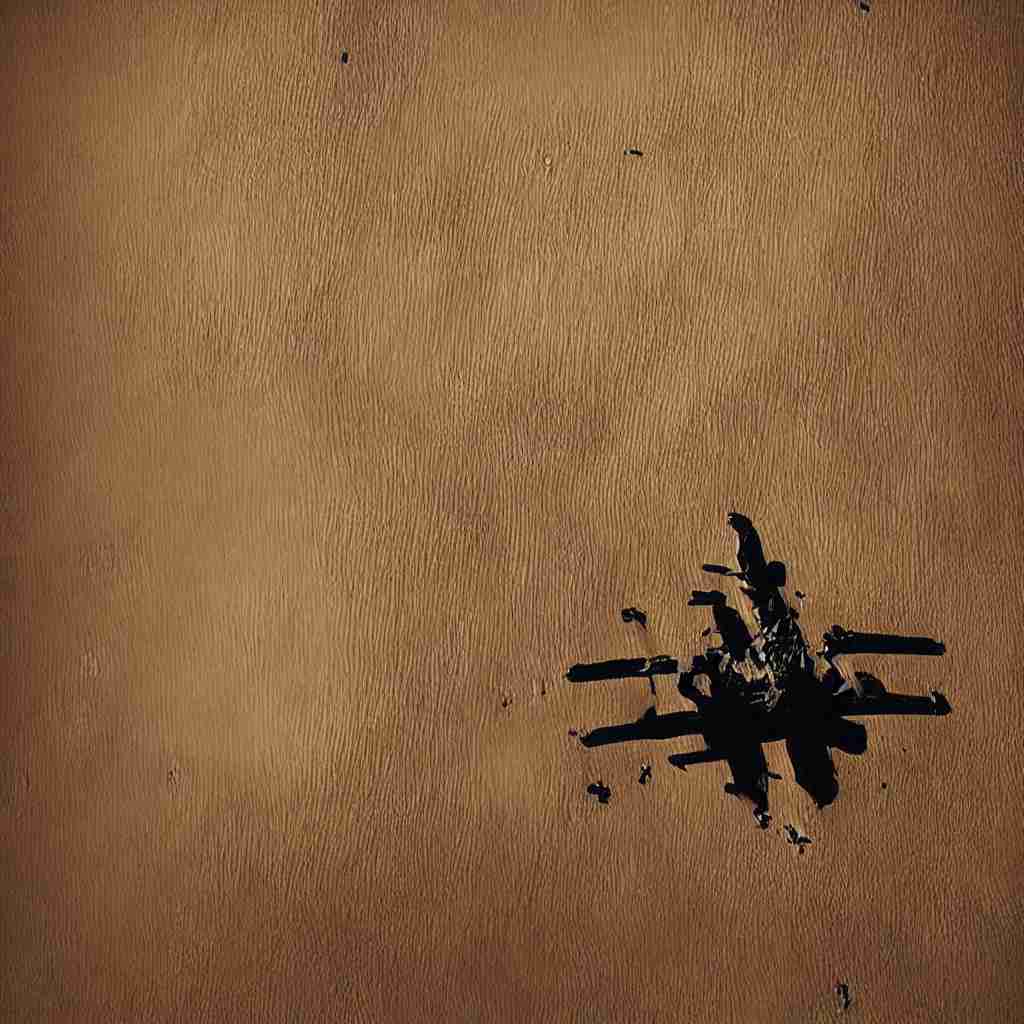}
    \end{minipage}
    \begin{minipage}{0.22\columnwidth}
        \includegraphics[width=\linewidth]{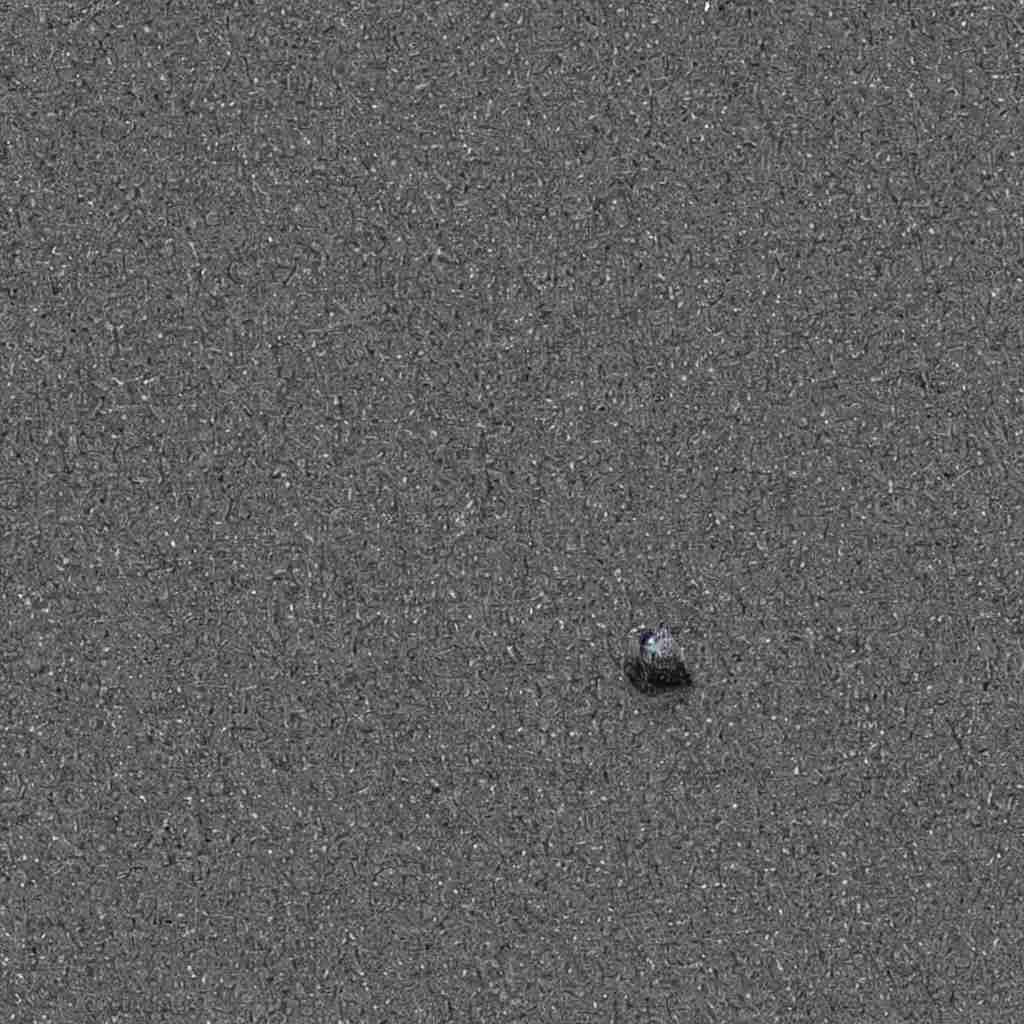}
    \end{minipage}
    \begin{minipage}{0.22\columnwidth}
        \includegraphics[width=\linewidth]{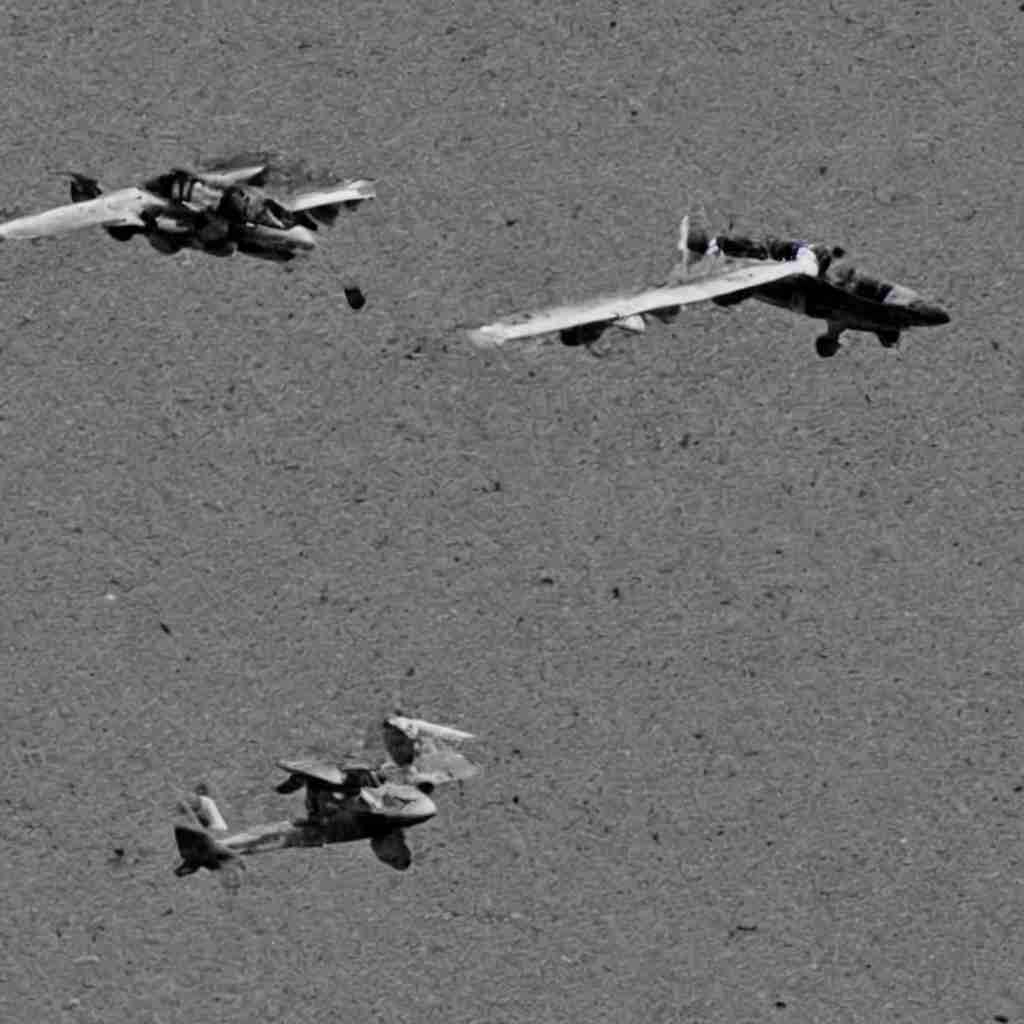}
    \end{minipage}
    \begin{minipage}{0.22\columnwidth}
        \includegraphics[width=\linewidth]{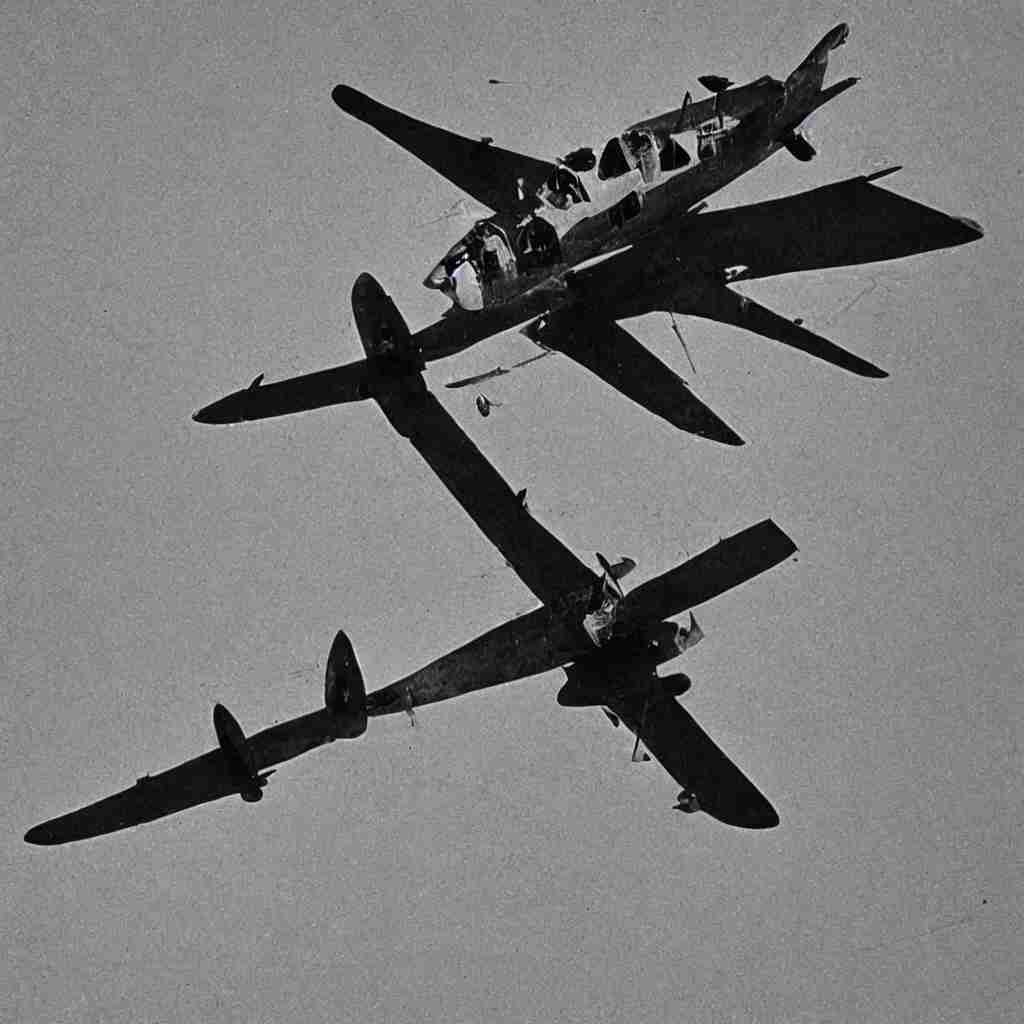}
    \end{minipage}
    \vspace{0.1cm} % Adjust this space as needed
    \caption{Failures from the trained model}
    \label{fig:fail_cases}
\end{figure}

In this setting, the output is influenced by the quality of the input dataset and the training stages. Due to computational constraints, advanced models like Stable Diffusion 2 and 3 were not explored in this study. The developed model has the capability to generate sonar images. However, it is uncertain how consistent the results are. In the experiments, it successfully produced images in 6 out of 10 trials, but sometimes the model hallucinates and leads to inaccurate responses. Some of such failure cases are shown in Fig.~\ref{fig:fail_cases}. Hence, in this pilot study, we develop a proof of concept rather than a comprehensive model.

\subsection{State-of-the-art Comparison}
\label{sec: sota comparison}

To the best of our knowledge, our work is the first work in text-conditioned image generation for underwatersonarimagery, and it lacks benchmark datasets, thereby hindering our ability to fully assess the model's performance. However, to showcase the efficacy of the proposed architecture, we compared our work with existing similar image-to-image synthesis work and tabulated the results in the Table.~\ref{tab: sota Image-to-image evaluation}. Our model outperforms the state-of-the-art models \cite{jmse11061103},
\cite{10534346},\cite{10453275} in terms of image quality, with high SSIM \& PSNR values and low FID value.

\begin{table}[h]
    \centering
    \begin{tabular}{|c|c|c|c|}
        \hline
        \textbf{Method} & \textbf{FID} & \textbf{SSIM} & \textbf{PSNR} \\
        \hline
        Zhiwei et al.\cite{jmse11061103} & 138.56 & 0.2512 & 11.1764 \\ 
        \hline
        Wen et al.\cite{10534346} & 0.2527 & - & - \\ 
        \hline
         Yang et al.\cite{10453275} & 147.6 & 0.37 & 16.3 \\ 
        \hline
        Synth-SONAR (Ours) & 3.8 & 0.381 & 12.730 \\ 
        \hline 
    \end{tabular}
    \caption{\textcolor{black}{State-of-the-art Comparison}}
    \vspace{.1cm}
    \label{tab: sota Image-to-image evaluation}
\end{table}

\vspace{-0.5cm}
\section{Conclusion}
\label{sec: conclusion}
In this work, we present ``Synth-SONAR", a novel framework for sonar image synthesis that leverages dual diffusion models and GPT prompting. Synth-SONAR overcomes traditional data collection challenges by creating a large, diverse, and high-quality sonar dataset. The framework uniquely integrates Generative AI-based style injection with real and simulated data and utilizes a dual text-conditioning diffusion model to generate synthetic sonar images from both high-level (coarse) and low-level (detailed) prompts. This approach bridges the gap between textual descriptions and sonar image generation, achieving state-of-the-art performance in enhancing image diversity and realism. Future work could focus on expanding Synth-SONAR across different underwater research applications.

\section{Acknowledgements}
\label{sec: acknowledgements}
This work was partially supported by the Naval Research Board (NRB), DRDO, Government of India under grant number: NRB/505/SG/22-23.

%%%%%%%%% REFERENCES
{\small
\bibliographystyle{ieee_fullname}
\bibliography{SYNTH-SONAR}
}

\end{document}